\pdfoutput=1

\PassOptionsToPackage{table}{xcolor}
\documentclass[11pt]{article}

\usepackage{acl}

\usepackage{booktabs}
\usepackage{multirow}
\usepackage{times}
\usepackage{latexsym}
\usepackage[table]{xcolor}
\usepackage{calc}             
\usepackage[most]{tcolorbox}  

\usepackage[T1]{fontenc}
\usepackage[most]{tcolorbox}
\usepackage{booktabs}
\usepackage{multirow}
\usepackage{xcolor} 
\usepackage[most]{tcolorbox}

\usepackage{tcolorbox}
\tcbset{fancyquote/.style={
  enhanced,
  colback=gray!5,
  colframe=black!30,
  boxrule=0.4pt,
  arc=2pt,
  left=6pt,
  right=6pt,
  top=6pt,
  bottom=6pt
}}

\newenvironment{fancyquote}{\begin{tcolorbox}[fancyquote]}{\end{tcolorbox}}

\usepackage[utf8]{inputenc}

\usepackage{microtype}

\usepackage{inconsolata}

\usepackage{graphicx}

\title{Telling Speculative Stories to Help Humans Imagine the Harms of Healthcare AI} 

\author{Xingmeng Zhao, Tongnian Wang, Dan Schumacher,\\ \textbf{Veronica Rammouz,} \and \textbf{Anthony Rios} \\
  The University of Texas at San Antonio\\
  \texttt{\{xingmeng.zhao, anthony.rios\}@utsa.edu} \\}

\begin{document}
\maketitle
\begin{abstract}
Artificial intelligence (AI) is rapidly transforming healthcare, enabling fast development of tools like stress monitors, wellness trackers, and mental health chatbots. However, rapid and low-barrier development can introduce risks of bias, privacy violations, and unequal access, especially when systems ignore real-world contexts and diverse user needs. Many recent methods use AI to detect risks automatically, but this can reduce human engagement in understanding how harms arise and who they affect. We present a human-centered framework that generates user stories and supports multi-agent discussions to help people think creatively about potential benefits and harms before deployment. In a user study, participants who read stories recognized a broader range of harms, distributing their responses more evenly across all 17 harm types. In contrast, those who did not read stories focused primarily on privacy and well-being (79.1\%). Our findings show that storytelling helped participants speculate about a broader range of harms and benefits and think more creatively about AI’s impact on users. Dataset and code are available at https:~\url{https://anonymous.4open.science/r/storytelling-healthcare/README.md}.

\end{abstract}

\section{Introduction}
Artificial intelligence (AI) is increasingly embedded in everyday domains such as finance, healthcare, and law~\cite{ashurst2020navigating}. In healthcare, AI tools including stress monitors~\cite{kargarandehkordi2025fusing}, wellness trackers~\cite{fabbrizio2023smart}, and mental health chatbots~\cite{macneill2024effectiveness} can directly affect users’ well-being. New prompting approaches such as vibe coding\cite{chow2025technology} allow non-experts to describe desired system behavior in natural language, enabling rapid prototyping of AI applications, for example through platforms like CareYaya\cite{vibecodingmedium2023}. However, these developments introduce risks related to fairness, bias, and accountability~\cite{weidinger2023sociotechnical}, which are especially critical in healthcare, where small errors can cause harm, including delayed treatment, privacy loss, and health inequities~\cite{roller2020open, chinta2025ai}. AI systems without appropriate safeguards may ultimately harm the users they aim to support~\cite{shelby2022identifying, shelby2023sociotechnical}. Although governments have begun responding through efforts such as the EU AI Act~\cite{europeanparliament2023AIAct} and a U.S. Executive Order~\cite{biden2023executive}, which emphasize transparency and accountability~\cite{khan2025towards}, regulation remains slow and fragmented, making early ethical foresight essential for aligning AI systems with human values~\cite{saxena2025ai}.
\begin{figure}[t]
    \centering
    \includegraphics[width=.55\linewidth]{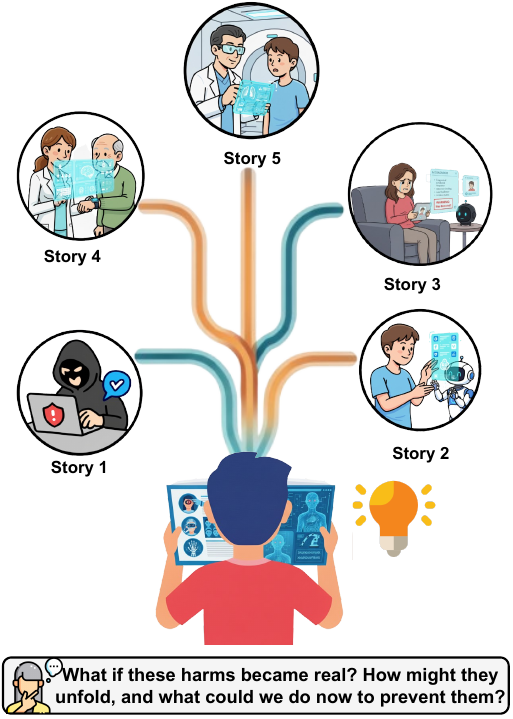}
    \caption{\textcolor{black}{Illustration of using speculative stories to help people imagine potential harms and benefits of healthcare AI and foster more creative and ethical thinking.}}
    \label{fig:introfig}
    \vspace{-1.7em}
\end{figure}
Ethical challenges in AI are commonly addressed through two complementary approaches: documenting known risks and anticipating potential harms early in design. Model cards~\cite{mitchell2019model} describe a system’s purpose, behavior, and limitations, and have been extended with interactive~\cite{crisan2022interactive} and structured formats~\cite{bhat2023aspirations}. Building on this work, RiskRAG~\cite{rao2025riskrag} automatically generates risk summaries from model cards and real-world incident data. While such automation helps scale ethical assessment, it may reduce opportunities for human reflection~\cite{kosmyna2025your} and introduce new harms~\cite{dutta2020there}. Another line of research focuses on anticipating misuse~\cite{herdel2024exploregen} and harm~\cite{deng2025supporting, saxena2025ai} early in the design process, using tools such as AHA!\cite{buccinca2023aha} and Farsight\cite{wang2024farsight}. However, as ethical reflection becomes increasingly automated, users may rely on AI judgment rather than their own, making ethical and contextual issues harder to recognize.
This challenge is particularly critical in healthcare, where small design mistakes can cause serious harm~\cite{mennella2024ethical, gilbert2025could}. Many risks only emerge after deployment, when AI systems operate in complex real-world settings~\cite{mun2024particip, kingsley2024investigating}. For example, mental health applications may fail to detect crises for certain populations, such as adolescents or non-native speakers~\cite{zhai2024effects}. Although existing tools often rely on automation to predict these risks, this can distance humans from ethical reasoning. Speculative design and design fiction offer an alternative by using imagined scenarios to explore how technologies might succeed or fail~\cite{rahwan2025science}. However, few approaches meaningfully support human participation in ethical foresight or integrate speculative thinking into real design workflows. Speculative storytelling addresses this gap by helping people reason about AI’s potential benefits and harms within realistic contexts~\cite{li2025one}.
Building on \citet{klassen2022run}, who use speculative fiction to examine emerging technologies, we apply storytelling to prompt early ethical reflection in AI design (Figure~\ref{fig:introfig}). Our approach tests whether stories encourage creative human speculation about potential benefits and harms, rather than relying on AI to anticipate risks. We introduce a human-centered framework that combines automated user story generation with structured red-team discussions. Unlike plot-planning methods~\cite{xie2024creating}, our approach generates context-sensitive stories grounded in users’ identities, behaviors, and needs. These stories help participants envision realistic success and failure scenarios, improving their ability to identify ethical and social risks. Using model cards as an evaluation tool, we show that story-driven discussions lead to more context-specific, detailed, and diverse expressions of potential harms.
Our contributions are twofold.
\textbf{(1)} We introduce a human-centered method that automatically generates context-sensitive user stories to help people imagine how an AI system could help or harm users before it is developed or deployed.
\textbf{(2)} We present a user study showing story-driven discussions with AI agents help participants explore potential risks and benefits more creatively and think more broadly about ethical issues.

\vspace{-1mm}
\section{Related Work}
\vspace{-1mm}
\noindent \textbf{Model Cards Framework.}
Model cards document an AI model’s purpose, performance, data sources, and limitations~\cite{mitchell2019model}. Later work improved their usability and scale: \citet{crisan2022interactive} created \textit{Interactive Model Cards} for exploring subgroup results, \citet{bhat2023aspirations} developed \textit{DocML} to guide non-experts, and \citet{rao2025riskrag} introduced \textit{RiskRAG}, which uses retrieval-augmented generation to summarize risks from model cards and incident reports. \citet{derczynski2023assessing} proposed \textit{Risk Cards} to describe failure cases in context. While these tools increase transparency, they rely on automation to fill ethical gaps, which can reduce opportunities for human reflection. Our approach instead uses AI to support human speculation, helping people imagine how systems might succeed or fail before deployment.

\vspace{1.5mm}
\noindent \textbf{Speculative Design.}
Speculative design uses imagined scenarios to explore how future technologies might affect people and society before they are built. Rather than predicting outcomes, it relies on what-if stories to prompt reflection on assumptions, values, and potential harms~\cite{klassen2022run, hoang2018can}, treating fiction as a tool for early ethical reasoning~\cite{rahwan2025science}. Prior work has applied this approach using AI-generated failure cases~\cite{buccinca2023aha}, risk prompts embedded in prototyping workflows~\cite{wang2024farsight}, and participatory methods, like Fiction Probes in healthcare~\cite{hoang2018can} and the Black Mirror Writers Room~\cite{klassen2022run}. Other extensions include participatory workshops, crowdsourced case studies, and AI-assisted red-teaming to broaden speculative exploration~\cite{mun2024particip, radharapu2023aart}, as well as non-narrative artifacts like generated comments or judgments~\cite{ballard2019judgment}. In contrast to directly generating harms, our approach uses AI-generated stories to prompt human reflection and help participants imagine harms.
\vspace{1.5mm}
\noindent \textbf{Language-Based World Modeling.}
Humans imagine situations to anticipate outcomes, explore alternatives, and guide decisions~\cite{addis2009constructive}. This ability relies on mental world modeling, in which people form internal representations of objects, events, and relationships to simulate possible futures~\cite{johnson1983mental}, supporting causal and counterfactual reasoning for planning and problem solving~\cite{lecun2022path}. Recent work shows large language models (LLMs) exhibit related capabilities through language. LLMs can act as text-based world models that simulate state transitions over time~\cite{xie2025making}, generate coherent and evolving environments in response to user actions~\cite{wang2023bytesized32}, and reason through internal multi-persona dialogue~\cite{wang2024unleashing}. Embodied agents further extend this idea by using internal world models to predict environments, infer user goals, and adapt to users’ mental models~\cite{fung2025embodied}. Building on this perspective, we frame story generation as language-based world imagination, where LLMs construct self-consistent narrative worlds to reason about possible futures and their social, ethical, and technical implications.

\begin{figure*}[t]
    \centering
    \includegraphics[width=.8\linewidth]{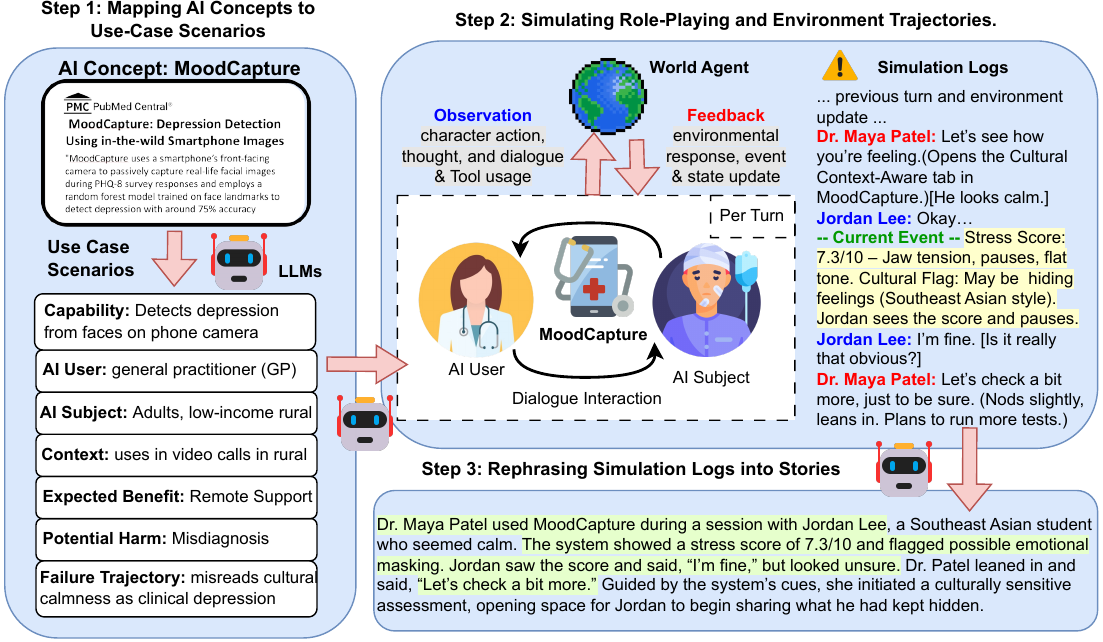}
    \caption{\textcolor{black}{Overview of the Storytelling Framework. We first generate use case scenarios from AI concepts sourced from PubMed, Wired, and industry app descriptions. Next, we simulate role-playing and environment trajectories for each scenario, producing detailed simulation logs. Finally, we rephrase these logs into short stories that illustrate both potential benefits and harms of the AI system.}}
    \vspace{-1.5em}
    \label{fig:storytelling}
\end{figure*}

\vspace{1.5mm}
\noindent \textbf{Automated Story Generation.} 
Early work on automated story generation emphasized explicit plot modeling, often drawing on narrative theories such as Propp’s functions to structure events, and typically adopted a two-stage pipeline that first planned key events and then realized them as full scenes~\cite{propp1968morphology, alhussain2021automatic}. With LLMs, this paradigm has shifted toward unified frameworks in which a single model jointly plans and writes narratives: Agents’ Room coordinates LLM-based character agents for collaborative story enactment~\cite{huotagents}, Dramatron decomposes screenplay generation into structured components such as loglines and dialogue~\cite{mirowski2023co}, and HOLLMWOOD uses LLM role-play to produce interactive, character-centered stories~\cite{chen2024hollmwood}. \citet{han2024ibsen} introduce a Director–Actor framework for interactive scriptwriting, which \citet{yu2025multi} extend with hierarchical role separation, while BookWorld adds a world agent to track global state and balance coherence with creativity~\cite{ran2025bookworld}.

\section{Methodology}
\vspace{-1mm}
In this section, we describe our prompting strategy for automated user story generation, as shown in Figure~\ref{fig:storytelling}. The goal is to speculate on both the benefits and potential risks of early AI diagnosis and decision making, imagining how they might help or cause harm in real-world use. First, we translated each AI concept into a realistic use case that defined its users, context, and intended purpose. Then, we simulated interactions between people, the AI system, and its environment to explore possible outcomes. Finally, we transformed these simulation logs into short stories that helped people reflect on future impacts and ethical implications.

\vspace{1.5mm}
\noindent \textbf{Step 1: Mapping AI Concepts to Use-Case Scenarios.}
We began by manually collecting 38 AI concepts in the consumer health domain. These examples were drawn from three sources: Wired articles, industry product descriptions, and PubMed research papers~\cite{saxena2025ai}~\footnote{The full list is available in our repository}. Each AI concept represented a potential consumer health application, such as estimating heart rate from smartphone camera input or monitoring mental well-being through daily behavior tracking. The set collected covered multiple domains, including mental health, chronic illness management, elderly care, and public health. We then used GPT-4o to generate structured model specifications for each concept, detailing the model name, task type, inference approach, and data requirements.

Next, we used each specification as input to generate a set of ethically sensitive use-case scenarios. Each use case was represented as a 7-tuple \(S = (a, u, s, x, b, h, f)\), where \(a\) denoted the AI’s capability, \(u\) the intended user (e.g., clinician), \(s\) the subject (e.g., patient), \(x\) the input or usage context, \(b\) the expected benefit, \(h\) the potential harm, and \(f\) the failure trajectory (e.g., possible unintended or problematic uses)~\cite{shaoprivacylens}. We used these structured representations to generate narrative user stories, which were then employed in red-teaming sessions to examine both user value and potential unintended harms. The full prompts used to extract model specifications and generate use cases are provided in Figure~\ref{sec:prompt-seed} in Appendix.

\vspace{1.5mm}
\noindent \textbf{Step 2: Simulating Role-Playing and Environment Trajectories.}
In this step, our system expanded each structured use case into detailed Role-Playing and Environment Trajectory logs that simulated how agents acted within an evolving world model. Our approach built on \textit{Solo Performance Prompting (SPP)}~\cite{wang2024unleashing}, a prompting technique in which a single LLM internally simulated multiple expert roles and engaged in self-collaboration within one prompt. This design enabled the model to construct an internal world model that supported multi-perspective reasoning and coherent simulation of role-based interactions.

We extended SPP by introducing a \textit{world agent}~\cite{ran2025bookworld}, a language-based simulator that maintained environmental coherence and handled non-dialogue interactions such as movement, tool use, or object manipulation. Each simulated role produced a structured output per turn consisting of three components:
(1) \textit{thoughts}, enclosed in brackets (e.g., [I need to know if the patient is under stress]), representing internal reasoning;
(2) \textit{actions}, enclosed in parentheses (e.g., (Dr. Patel opens the cultural assessment tab)), representing observable behavior; and
(3) \textit{dialogue}, written in plain text, representing spoken communication.
This structure allowed the world agent to separate internal reasoning from external actions and update the environment accordingly. When an action affected the world, such as retrieving patient data, adjusting a protocol, or activating a sensor, the agent simulated the corresponding system response. In effect, the model operated as a language-based world simulator, incrementally constructing an evolving narrative environment through agent–environment interaction. For example, a doctor agent might issue the following action:
\begin{quote}
\texttt{(Dr. Patel opens the Cultural Context-Aware Assessment tab)}
\end{quote}
The world agent interpreted this as an interaction with a virtual diagnostic tool. It considered the current session context (e.g., a teletherapy consultation), relevant background knowledge (e.g., cultural models of stress expression), and prior AI-generated alerts to simulate the tool’s response. The resulting output might appear as follows:
\begin{quote}
\texttt{– Current Event –}
\texttt{Stress Score: 7.3/10 – Detected jaw tension, micro-pauses, and flat vocal tone.}
\texttt{Cultural Flag: Possible emotional masking (Southeast Asian expression style).}
\texttt{Jordan saw the score on screen and became slightly hesitant.}
\end{quote}
The response was returned to the doctor role and informed their next move, whether a reply, a new question, or a follow-up action. The world agent then updated the simulation state by adjusting variables such as the patient’s emotional profile or the alert level. These updates maintained coherence and allowed role behavior to evolve naturally with the unfolding context. This step produced a log capturing the full trajectory of the simulation, including role thoughts, dialogue, actions, tool calls, and resulting environment changes. This log served as the basis for generating the evolving narrative in the next step. Prompt is provided in Figure~\ref{sec:prompt-logs}.

\vspace{1.5mm}
\noindent \textbf{Step 3: Rephrasing Simulation Logs into Stories.} After the simulation, the system collected logs from Step 2 and prompted an LLM to rephrase them into a concise, five-sentence narrative. This step transformed structured logs into stories that preserved the main events, role dynamics, and emotional flow of the interaction. The full rephrasing prompt is provided in Figure~\ref{sec:rephrase} in the Appendix.

\section{Experiments}
This section details our story generation datasets, evaluation metrics, and results.

\vspace{1.5mm}
\noindent \textbf{Dataset.} We used GPT-4o to generate ethically sensitive use-case scenarios from 38 consumer health AI solutions sourced from \textit{Wired}, industry product documentation, and PubMed. Each scenario acted as a narrative seed for simulation. For each AI concepts, we generated ten variations spanning different user roles (e.g., doctor, nurse, caregiver), settings (e.g., rural clinic, hospital, home), patient profiles (e.g., adolescent, older adult, multicultural family), and contextual conditions.

\vspace{1.5mm}
\noindent \textbf{Baseline.} 
We compared our method with a traditional plot-planning approach, where the model first outlines a plot before writing the story~\cite{yao2019plan, xie2024creating}. Using the same ethically sensitive seed, the baseline generated each story in a single step following a structured template. Each story consisted of five sentences designed to prompt ethical reflection. The template directs the LLM to describe the AI system’s purpose, the people involved, the everyday use context, potential ethical risks, and how user identity may influence harm or misinterpretation. The full baseline prompt is provided in Figure~\ref{sec:baseline} in Appendix.

\vspace{1.5mm}
\noindent \textbf{Setting for Pairwise Comparison.} We followed the evaluation setup from~\cite{li2025storyteller} to assess story quality across multiple dimensions. Stories were evaluated according to five criteria: \textbf{Creativity}, measuring the originality and imagination of the plot and characters; \textbf{Coherence}, assessing narrative clarity and logical flow; \textbf{Engagement}, capturing how well the story maintains reader interest; \textbf{Relevance}, measuring consistency with the given prompt or scenario; and \textbf{Likelihood of Harm or Benefit}, evaluating whether the story depicts realistic AI behavior with meaningful social consequences. Following the arena-hard-auto evaluation method~\cite{licrowdsourced}, we used stories generated with the story-planning approach (by GPT-4o) as the reference baseline and compared them with stories produced by our method across different LLMs. For each metric, GPT-4o or human judges determined which story performs better or mark them as indistinguishable (``Tie''). Win rates were calculated based on these pairwise preferences, and the full configuration details and evaluation prompts are included in the Appendix. To eliminate positional bias, we randomized the order of story pairs and alternate their positions across comparisons. See Figures~\ref{sec:criteira1} and~\ref{sec:criteria2} for detailed criteria in Appendix.

\begin{table*}[t]
\centering
\resizebox{.9\linewidth}{!}{
\begin{tabular}{llrrrrrr}
\toprule
\textbf{Story Type} & \textbf{Model} & \textbf{Creativity} & \textbf{Coherence} & \textbf{Engagement} & \textbf{Relevance} & \textbf{Likelihood} & \textbf{Overall (Avg)} \\
\midrule
\multirow{3}{*}{Baseline}
  & GPT4o   & 50.00 & 50.00 & 50.00 & 50.00 & 50.00 & 50.00 \\
  & Llama3  & 59.25 & 71.55 & 76.15 & 71.60 & 70.00 & 69.71 \\
  & Gemma   & 65.25 & 68.30 & 80.15 & 71.20 & 78.90 & 72.76 \\
\midrule
\multirow{3}{*}{Storytelling (ours)}
  & GPT4o   & \textbf{63.15} & \textbf{63.45} & \textbf{59.35} & \textbf{70.90} & \textbf{69.10} & \textbf{65.19} \\
  & Llama3  & \textbf{79.50} & \textbf{94.75} & \textbf{89.45} & \textbf{85.65} & \textbf{96.85} & \textbf{89.24} \\
  & Gemma   & \textbf{89.45} & \textbf{92.15} & \textbf{92.75} & \textbf{85.65} & \textbf{96.05} & \textbf{91.21} \\
\midrule
\multirow{3}{*}{w/o Environment Trajectories}
  & GPT4o   & 24.35 & 21.20 & 26.60 & 21.05 & 18.85 & 22.41 \\
  & Llama3  & 31.20 & 52.70 & 39.20 & 51.75 & 50.85 & 45.14 \\
  & Gemma   & 55.30 & 74.35 & 78.80 & 73.45 & 85.50 & 73.48 \\
\midrule
\multirow{3}{*}{w/o Role-Playing}
  & GPT4o   & 18.05 & 45.80 & 47.90 & 49.70 & 48.30 & 41.95 \\
  & Llama3  & 49.35 & 73.65 & 72.00 & 73.70 & 74.35 & 68.61 \\
  & Gemma   & 79.45 & 86.80 & 83.95 & 83.15 & 91.05 & 84.88 \\
\bottomrule
\end{tabular}}
\caption{Overall results of different models and methods. \textbf{Storytelling (ours)} achieves the best performance across all metrics. Values denote win rates (\%). The highest score for each model is in \textbf{bold}. ``w/o Environment Modeling'' means the model performs only role-playing without modeling event progress, and ``w/o Role-Playing'' means it predicts sequential events without character dialogue.}
\label{tab:story_win_by_type_updated}
\vspace{-1em}
\end{table*}

\vspace{1.5mm}
\noindent \textbf{LLM-as-a-Judge Evaluation.} As shown in Table~\ref{tab:story_win_by_type_updated}, our Storytelling method consistently outperforms all baselines across every metric. When combined with the Gemma model, it achieves win rates of 89.45\% for creativity, 92.15\% for coherence, 92.75\% for engagement, 85.65\% for relevance, and 96.05\% for likelihood, yielding an overall average of 91.21\%. In contrast, the baseline Gemma records 72.76\%, and Llama3 reaches 69.71\%, indicating gains of roughly +15–25 points across dimensions. We use GPT-4o as the evaluator with the temperature set to 0.1 to ensure deterministic and consistent judgments across comparisons. Comparing models, baseline Gemma slightly outperforms Llama3 in most metrics, but under the Storytelling framework, Llama3 nearly closes the gap with an overall score of 89.24\%. Interestingly, Llama3 surpasses Gemma in coherence (94.75 vs. 92.15) and performs equally well in relevance (85.65), suggesting that Llama3 shows stronger structural reasoning and coherence, while Gemma excels in narrative creativity and expressiveness. Overall, these results demonstrate that integrating world and role-based modeling enables models to reason about events, sustain coherent narratives, and produce stories that are both imaginative and believable. For robustness, we report results from two additional LLM-as-a-judge models in the appendix~\ref{app:additional_judges}.

\begin{figure}[t]
    \centering
    \includegraphics[width=\linewidth]{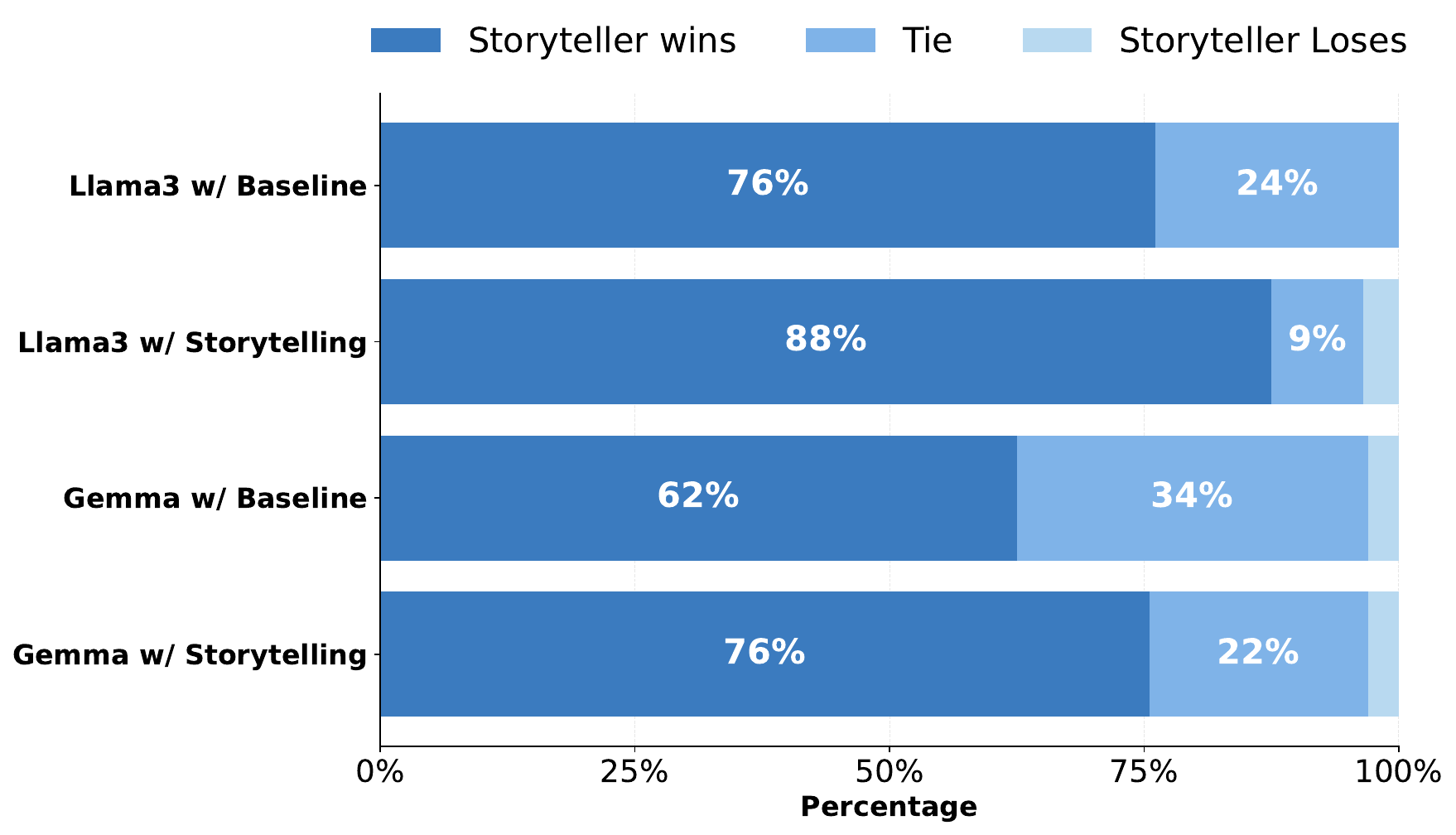}
    \caption{\textcolor{black}{Results of human preference evaluation. Our Storytelling method achieves strong preference wins against the baseline, with 88\% preference using Llama3 and 76\% using Gemma3.}}
    \label{fig:future}
    \vspace{-1em}
\end{figure}

\vspace{1.5mm}
\noindent \textbf{Human Evaluation.} To complement the LLM-as-a-Judge evaluations and reduce potential bias, we conducted human preference evaluations. Two graduate student annotators independently evaluated 100 story pairs for each model and method. As shown in Figure~\ref{fig:future}, our \textbf{Storytelling} method is consistently preferred over all baselines, achieving 88\% preference for Llama3 and 76\% for Gemma, patterns that align with GPT-4o based evaluations. Notably, human judges show a slightly stronger preference for Llama3, suggesting it produces stories that are easier to follow and more engaging, while Gemma tends to generate more expressive and stylistically rich narratives. We further measure inter-annotator consistency using Cohen’s kappa~\cite{cohen1960coefficient}. As shown in Table~\ref{tab:kappa}, agreement scores range from 0.619 to 0.729 across models and methods, indicating substantial reliability. Overall, both human and LLM evaluations consistently agree that Storytelling outperforms all baselines, and differences between Gemma and Llama3 reflect LLM preference for detail versus human preference for clarity.

\begin{table*}[t]
\centering
\resizebox{.9\linewidth}{!}{
\begin{tabular}{llrrrrrrr}
\toprule
\textbf{Method} & \textbf{Model} & \multicolumn{4}{c}{\textbf{DistinctL-n}} & \multicolumn{2}{c}{\textbf{Diverse}} \\
\cmidrule(lr){3-6} \cmidrule(lr){7-8}
& & \textbf{DistinctL-2} & \textbf{DistinctL-3} & \textbf{DistinctL-4} & \textbf{DistinctL-5} & \textbf{Verbs} & \textbf{Avg Word Count} \\
\midrule
\multirow{3}{*}{Baseline}
         & GPT4o   & 5.692 & 5.794 & 5.798 & 5.799 & 0.984 & 122 \\
         & Llama3  & 5.728 & 5.820 & 5.951 & 5.961 & 0.934 & 175 \\
         & Gemma   & 5.837 & 5.939 & 5.946 & 5.946 & \textbf{0.979} & 141 \\
\midrule
\multirow{3}{*}{Storytelling (ours)}
         & GPT4o   & 5.696 & \textbf{5.818} & \textbf{5.824} & \textbf{5.825} & 0.978 & 125 \\
         & Llama3  & \textbf{5.840} & \textbf{6.104} & \textbf{6.158} & \textbf{6.174} & 0.937 & 179 \\
         & Gemma   & \textbf{5.863} & \textbf{6.042} & \textbf{6.062} & \textbf{6.065} & 0.955 & 159 \\
\midrule
\multirow{3}{*}{w/o Environment Trajectories}
         & GPT4o   & 5.585 & 5.687 & 5.693 & 5.693 & 0.974 & 110 \\
         & Llama3  & 5.745 & 5.980 & 6.025 & 6.036 & \textbf{0.953} & 155 \\
         & Gemma   & 5.734 & 5.861 & 5.873 & 5.873 & 0.978 & 131 \\
\midrule
\multirow{3}{*}{w/o Role-Playing}
         & GPT4o   & \textbf{5.698} & 5.789 & 5.794 & 5.794 & \textbf{0.988} & 121 \\
         & Llama3  & 5.722 & 5.819 & 5.849 & 5.858 & 0.936 & 175 \\
         & Gemma   & 5.834 & 5.935 & 5.942 & 5.943 & 0.977 & 141 \\
\bottomrule
\end{tabular}}
\caption{Diversity results of different models and methods. We report DistinctL-2 through DistinctL-5 (higher is more diverse), Diverse Verbs, and the average story length. The highest score for each model is highlighted in \textbf{bold}.}
\label{tab:avg_distinct_diverse}
\vspace{-1.3em}
\end{table*}

\vspace{1.5mm}
\noindent \textbf{Ablation Study.} To evaluate each component’s contribution, we performed two ablations by removing the role-playing or environment trajectory mechanisms. As shown in Table~\ref{tab:story_win_by_type_updated}, removing environment trajectory, where the model performs only role-playing without predicting how events evolve, produced the largest drop across all models. For Gemma, coherence dropped by 17.7 and relevance by 12.2, showing that modeling event progression is vital for narrative logic. Removing role-playing, which limits the model to sequential event prediction without character perspectives, reduced creativity (–10.0) and engagement (–8.8). Overall, environment trajectory maintains coherent story flow, while role-playing adds diversity and emotional depth, making both essential for effective story generation. See the repository for the full ablation prompt template.

\vspace{1.5mm} 
\noindent \textbf{Automated Evaluation.} Additionally, we assess story diversity using DistinctL-n~\cite{li-etal-2016-diversity} and Diverse Verbs~\cite{fan2019strategies}, which measure lexical variety and action diversity, more details can be found in Appendix. As shown in Table~\ref{tab:avg_distinct_diverse}, our Storytelling method achieves consistently higher diversity than the baselines. With Llama3, it reaches the highest scores on DistinctL-3 to DistinctL-5 (6.104, 6.158, and 6.174), indicating richer and less repetitive text. Gemma also shows steady improvements, achieving 5.863 on DistinctL-2 and maintaining strong overall diversity. The environment trajectory ablation attains the highest Diverse Verbs score (0.988) but lower DistinctL-n, suggesting a balance between lexical variety and action diversity. Overall, our method generates more detailed and varied narratives while preserving structural consistency.

\section{User Study}
We conducted a user study to examine whether engaging with benefit and harm stories enhances participants’ ability to speculate about the impacts of AI systems. Rather than relying on AI-generated ideas, our goal is to prompt participants to actively reflect on potential risks and benefits. 
We assess this by evaluating how participants reason about these aspects when completing a speculative model card. All procedures were approved by our Institutional Review Board (IRB).


\vspace{1.5mm}
\noindent \textbf{Speculative Model Card Task.}
This study used a between-subjects design~\cite{mackenzie2016empirical}. Participants completed a speculative model card, a structured template describing an AI system’s intended use, benefits, and potential harms, under one of three conditions. In the \textsc{Control} condition, participants completed the model card directly. In the \textsc{Story-Only} condition, participants first read text-based benefit and harm stories, without discussion, before completing the model card. In the \textsc{Story} condition, participants engaged in a red-team discussion on our platform to explore benefit and harm stories before completing the same model card. The model card template is shown in Figure~\ref{fig:model-inter} in the appendix.

\vspace{1.5mm}
\noindent \textbf{User Study Results.} 
We conducted a user study with 45 participants to examine how storytelling-based discussions influence ethical reasoning in AI documentation. Participants completed a speculative model card task under three conditions: a control group that worked individually, a story-only group that read the text-based stories without discussion, and a treatment group that used our \textit{Story-Driven Red-Team Discussion Room}. The discussion platform enabled participants to engage with simulated expert personas in guided, story-based conversations about the potential benefits and harms of AI systems. Each session included three stages: a pre-survey, the model card completion task, and a post-survey evaluating perceived usefulness, trust, and engagement. We analyzed participants’ model card responses (benefit and harm use cases) and post-survey feedback to assess how narrative interaction supported ethical reflection. As an exploratory qualitative study, we focus on recurring themes rather than statistical power, and prior work shows that small samples are sufficient to reach thematic saturation, where few new themes emerge with additional data~\cite{hennink2022sample}. Results are organized into three key areas: (1) identifying potential benefits, (2) uncovering possible harms, and (3) linking harms to participants’ personal needs and contexts. We applied qualitative coding to classify harm and benefit types, with two annotators achieving moderate agreement (Cohen’s $\kappa = 0.4368$ for harms and $\kappa = 0.3968$ for benefits). Study design and full procedures are provided in Appendix~\ref{sec:userstudy}. As a robustness check, we conducted an LLM-based simulated survey under the same conditions, as described in Appendix~\ref{app:llm_survey}.

\vspace{1.5mm}
\noindent \textbf{Does Storytelling Help Identify More Harms?}
We analyzed responses across 17 harm subtypes defined by~\citet{shelby2023sociotechnical}. See Table~\ref{tab:consumer_health_ai_harms} in the appendix for the full category list. As shown in Table~\ref{tab:harm_subtypes}, the \textsc{Control} group concentrated on a small set of categories, primarily \textit{diminished health or well-being} (32.3\%), \textit{service or benefit loss} (24.2\%), and \textit{privacy violations} (22.6\%), which together accounted for the majority of reported harms. In contrast, both the \textsc{Story-only} and \textsc{Story} conditions exhibited more distributed coverage across harm types, with the \textsc{Story} condition showing the widest range of subtypes and lower concentration in any single category. Several harm categories, including \textit{cultural harms}, \textit{political and civic harms}, and \textit{tech-facilitated violence}, appeared only in the \textsc{Story} condition, suggesting that interactive narrative discussion supported recognition of less obvious and context-dependent harms. We quantified these differences using Shannon entropy ($H$), which measures distributional diversity across harm types. As shown in Table~\ref{tab:harm_subtypes}, entropy increased from 2.329 in the \textsc{Control} condition to 2.927 in the \textsc{Story-only} condition and to 3.701 in the \textsc{Story} condition. Bootstrap $t$-tests confirmed that both \textsc{Story-only} and \textsc{Story} exhibited significantly higher entropy than \textsc{Control} ($p<.001$), and that \textsc{Story} also showed significantly higher entropy than \textsc{Story-only} ($p<.001$). These results indicate that storytelling-based engagement, particularly interactive discussion, broadened participants’ awareness of potential harms and supported more diverse ethical reasoning.

\vspace{1.5mm}
\noindent \textbf{Does Storytelling Help Reveal More Benefits?}
We examined whether storytelling broadened participants’ recognition of potential benefits across 18 predefined subtypes, summarized from prior consumer health AI research~\cite{pedroso2025leveraging, chustecki2024benefits}. Detailed category descriptions are provided in Table~\ref{tab:consumer_health_ai_benefits} in the appendix. As shown in Table~\ref{tab:benefit_subtypes}, participants in the \textsc{Control} group concentrated on a small set of benefits, primarily \textit{decision support \& diagnostic augmentation} (25.4\%), \textit{continuous monitoring \& self-care} (23.8\%), and \textit{early detection \& prediction} (22.2\%), which together accounted for the majority of responses. In contrast, both the \textsc{Story-only} and \textsc{Story} conditions exhibited broader coverage across benefit types, with the \textsc{Story} condition showing the most diverse distribution and lower concentration in any single category. Several benefit subtypes, including \textit{accessibility \& disability support}, \textit{clinician workload relief}, and \textit{transparency \& trust}, appeared only in the \textsc{Story} condition, suggesting that interactive narrative discussion supported recognition of less salient or less frequently considered benefits.

We quantified these differences using Shannon entropy ($H$), which captures the evenness of the benefit distribution. As shown in Table~\ref{tab:benefit_subtypes}, entropy increased from 2.407 in the \textsc{Control} condition to 3.242 in the \textsc{Story-only} condition and to 3.868 in the \textsc{Story} condition. Bootstrap $t$-tests confirmed that both \textsc{Story-only} and \textsc{Story} exhibited significantly higher entropy than \textsc{Control} ($p<.001$), and that \textsc{Story} also showed significantly higher entropy than \textsc{Story-only} ($p<.001$). These results indicate that storytelling-based engagement—particularly interactive discussion—encouraged participants to recognize a more diverse and balanced set of potential AI benefits.

\vspace{1.5mm}
\noindent \textbf{What do People Say About Ethical Reflection in Speculative AI Documentation?} Post-survey responses indicated that Human–AI storytelling discussions fostered deeper ethical and contextual reflection on AI systems. Participants reported the narrative format helped them articulate risks that were otherwise difficult to express. For example, P9 shared that \textit{``It helped me to understand more,''} and P7 noted that \textit{``The story provides a concrete example of how AI can be harmful.''} Engaging with concrete narrative scenarios led participants to verbalize their reasoning about model risks in a think-aloud manner, supporting ethical reflection without requiring prior expertise. As P4 explained, \textit{``I could not think of [risks] really, but the story shifted my focus to the negative aspect of things which we usually ignore.''} Others observed that stories surfaced overlooked issues, such as \textit{``the lack of cultural context''} (P6) or emotional harms like \textit{``masking of feelings''} (P3), suggesting that narrative prompts helped surface subtle sociotechnical risks often missing from formal documentation.

Finally, participants found the storytelling approach both engaging and accessible. By embedding risk exploration within narrative contexts, the format allowed learners to focus on ethical reflection rather than technical complexity. As P12 remarked, \textit{``It makes the model more interesting and understandable,''} and P8 noted that the story \textit{``helped me to know how to use the AI tool,''} indicating that minimal prior expertise was required to engage meaningfully with ethical scenarios. These findings suggest that Human-AI storytelling discussions can sustain interest while supporting active, reflective engagement with model risks and benefit.

\section{Conclusion}
In this paper, we explored speculative storytelling as a method to improving human ability to anticipate both the benefits and risks of AI-driven healthcare systems before they are developed or deployed. By simulating realistic scenarios, this approach encourages critical reflection on how AI might succeed or fail, shifting safety evaluation from a reactive to a proactive process. Our findings show that storytelling improves people’s ability to anticipate how AI systems might help or harm in practice, highlighting the importance of human judgment over automated speculation in ethical evaluation.

\section{Limitation}
This work has several limitations that indicate directions for future extension rather than weaknesses. Our scenarios focus on consumer health and do not include regulated domains such as clinical decision support, finance, or law. While the framework could be applied to these areas, we have not yet tested it there. The scenarios are synthetic and derived from AI concepts with assistance from LLMs, enabling early exploration of ethical issues but not substituting for analysis of deployed systems. We intentionally use low-stakes, synthetic consumer-health scenarios and university participants to validate the storytelling method before deployment in regulated or clinical environments. This study should therefore be understood as a first step that demonstrates methodological feasibility rather than direct applicability to clinicians or patients, which we leave to future work.

We rely on a single LLM as a judge for pairwise comparisons. A single judge may favor certain writing styles or phrasing. To mitigate this, we randomize prompt order and report human agreement, but larger evaluations with multiple models would offer stronger validation. Our user study is small and includes mostly participants with technical backgrounds. The findings may not generalize to clinicians, patients, or policymakers, and we measure only short-term reflection rather than long-term impact.

The simulated expert discussions use predefined personas instead of real experts. This choice enables rapid iteration, but does not capture the full range of stakeholder perspectives. Our metrics (e.g., creativity, coherence, engagement, relevance, and likelihood of harm or benefit) are useful indicators but do not represent the groundtruth in safety. Finally, although we release code, prompts, and configurations, some results rely on proprietary APIs, which may change over time and limit exact reproducibility.

Overall, these limitations reflect practical design decisions for early-stage exploration of AI storytelling as a method for surfacing ethical risks. They suggest next steps in evaluating across domains, with larger and more diverse human studies, and with multiple evaluation models.

\section{Ethics Statement}
This study uses fictional stories to explore how people reason about potential risks and benefits of future AI systems in health contexts. The scenarios describe speculative technologies that do not currently exist. We clearly framed every story as hypothetical and avoided making claims about real clinical products or patient outcomes. This follows ARR and ACL guidance on disclosing potential societal effects while separating speculation from evidence.

Even with fictional framing, generated stories can reproduce bias or misleading claims. We reviewed outputs and removed content that could confuse readers or reinforce harmful stereotypes. These safeguards align with ACL ethics guidance related to fairness, sensitive attributes, and downstream harm. We present narrative outputs as prompts for reflection, not as predictions or endorsements.

Because stories can shape how readers think about AI, speculative harms and benefits must be contextualized. Prior ACL work shows that ethical sections should identify affected groups, describe potential harms, and discuss mitigation steps. We therefore state the audience and limits of interpretation, and we report findings in aggregate without making policy or clinical claims.

All human-subject activities were reviewed and approved by our Institutional Review Board (IRB). Participants provided informed consent and were compensated for their time. No identifying information was collected. These practices follow ethical norms for human studies referenced in ARR materials and broader research ethics standards.

We used existing large language models accessed through public APIs and did not retrain or fine-tune them. We release prompts and study materials to support transparency and allow others to audit or adapt the procedure. ACL guidance highlights documentation and reproducibility when model behavior may carry societal impact.

Finally, we acknowledge community expectations for proactive ethics communication. Tutorials and position work in the ACL community encourage explicit articulation of risks, stakeholders, and mitigation strategies, and support structured processes that help authors consider ethics early in system design. Our study aligns with these goals by examining how narrative framing can facilitate ethical reflection in the early stages of AI development.

\bibliography{custom}

\appendix


\section{Appendix}
\subsection{Additional Related Work}
\noindent \textbf{AI in Consumer Health.} AI tools in consumer health, like mobile apps, wearables, and telemedicine, help people manage chronic conditions and wellness (e.g., diabetes apps, fitness trackers)~\cite{ashurst2020navigating, triantafyllidis2019features, asan2023artificial}. A recent review found that 65\% of these tools are mobile apps, 25\% are robotics, and 10\% are telemedicine, mostly focused on personalized care and better outcomes~\cite{asan2023artificial}. Although many users find these tools helpful and easy to use, some remain hesitant to trust them in the absence of clear medical evidence or transparency around data use~\cite{oudbier2025patients}. Unregulated apps, such as mental health bots and symptom checkers, have grown faster than oversight, raising safety and fairness issues~\citealp{herpertz2025developing, freyer2024regulatory}. To address these gaps, researchers advocate early co‐design with patients and caregivers, co‐developing ethical checklists and participatory guidelines to surface hidden biases and workflow mismatches~\cite{madaio2020co, shi2025mapping}. They also suggest using ``AI Nutrition Labels'' to transparently communicate intended use, data sources and known limitations to end users~\cite{rozenblit2025towards, wachter2021fairness}.

\vspace{1.5mm}
\noindent \textbf{Ethical Harm in Healthcare and Well‑being.} AI in health can cause real risks, like biased decisions, unfair access, or unsafe advice~\cite{shelby2022identifying, shelby2023sociotechnical}. Addressing these issues early is essential~\cite{saxena2025ai, callahan2024standing}. One early check is the ``What \& Why'' assessment: does the AI solve a real healthcare need, how will its output be used, and what impact will it have~\cite{callahan2024standing}? \citet{saxena2025ai} propose the \textit{AI Mismatches} framework to identify gaps between a model’s actual performance and real-world user needs. \citet{li2025one} stresses the need for models to adapt across users and settings to avoid context-sensitive failures. To address evaluation blind spots, red-teaming clinical LLMs helps catch safety, privacy, and bias issues that standard tests miss~\cite{chang2025red}. Similarly, tools like the Health Equity Evaluation Toolbox use adversarial data to reveal demographic bias~\cite{pfohl2024toolbox}. 

\subsection{Case Study}
\begin{figure*}[t]
    \centering
    \includegraphics[width=.85\linewidth]{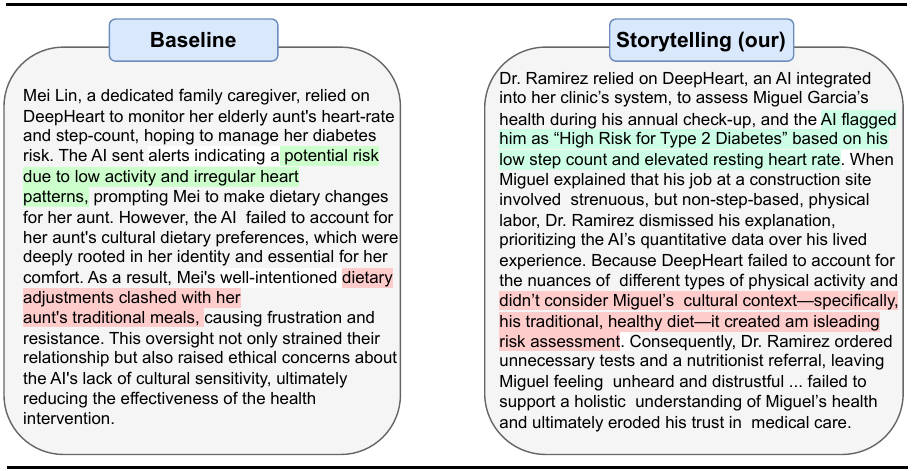}
    \caption{\textcolor{black}{A qualitative example showing how our storytelling method makes the AI’s decision process and its consequences easy to follow. Unlike a simple narrative description, the story explicitly surfaces what changed, why it changed, and how stakeholders were affected.}}
    \vspace{-1em}
    \label{fig:story1}
\end{figure*}

\begin{figure*}[t]
    \centering
    \includegraphics[width=.85\linewidth]{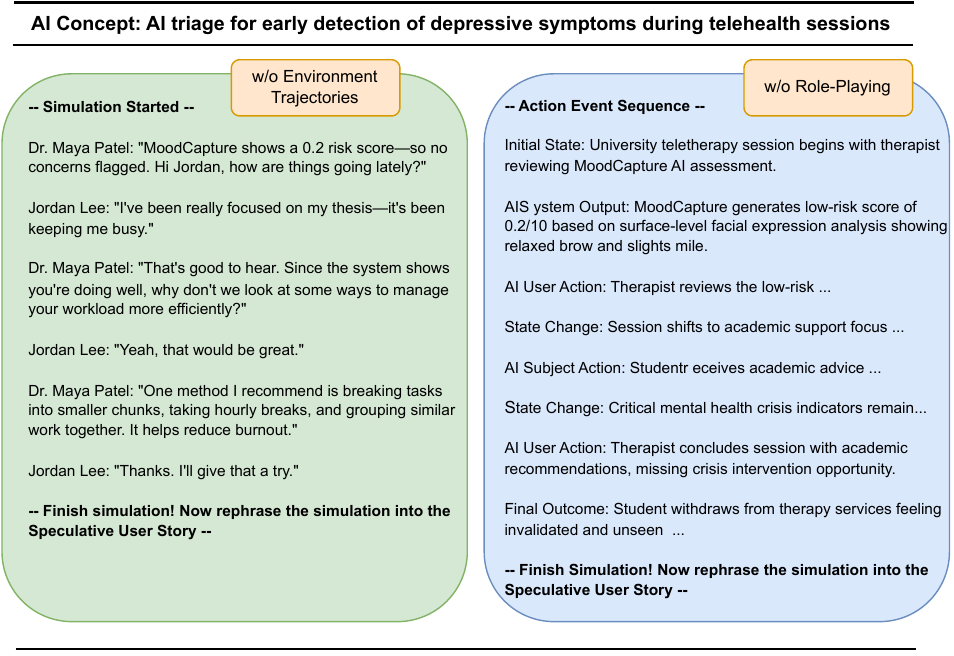}
    \caption{\textcolor{black}{A comparison of simulation logs under different ablations (w/o Environment Trajectories and w/o Role-Playing) to show the contribution of each component.}}
    \vspace{-1em}
    \label{fig:story2}
\end{figure*}

We conduct a qualitative analysis to understand how different storytelling configurations influence readers’ ability to reason about AI behavior. Our goal is not just to produce coherent narratives, but to evaluate whether a story actively helps readers see what happened, understand why it happened, and recognize its consequences. We therefore analyze whether the narrative (1) makes both positive and negative system outcomes visible, (2) clearly exposes the decision process that leads to those outcomes, and (3) encourages causal reasoning rather than surface-level emotional reactions (e.g., “AI is good” or “AI is dangerous”). When these elements are present, the story functions as an interpretive lens rather than a simple anecdote. Figure~\ref{fig:story1} provides an example where our method achieves this explanatory quality. To isolate the contribution of each narrative component, Figure~\ref{fig:story2} presents an ablation comparison, demonstrating that removing Environment Trajectories or Role-Playing reduces the richness of causal cues and results in flatter, less informative character behavior.


\subsection{Alignment Between Human and LLM-Based Evaluations}
Our study compares human preferences with LLM-as-a-judge at the overall method level. As reported in Section 4, human annotators strongly prefer our Storytelling method (88\% for Llama3 and 76\% for Gemma). This preference closely aligns with results obtained using GPT-4o as an automatic judge. Both humans and LLM judges reach the same conclusion: our method produces higher-quality stories than the baseline and ablation variants. We also observe consistent stylistic patterns across evaluators. Human annotators slightly prefer Llama3 because its stories are easier to follow, whereas Gemma tends to generate more expressive and detail-dense narratives, a distinction that is also reflected in GPT-4o’s scores.

A more fine-grained comparison between human and LLM judges at the level of individual dimensions (e.g., creativity, coherence, engagement) could provide additional insight. We did not collect dimension-specific human ratings because doing so would impose substantial cognitive load on annotators in this early-stage study, which was designed to focus on overall story quality. We will clarify this design choice in the revision and identify dimension-level human–LLM agreement as an important direction for future work. Overall, our results show strong alignment between human judgments and LLM judges in terms of global ranking and observed stylistic tendencies.

Below, we provide a qualitative illustration that helps explain the small differences between human and LLM preferences.

\vspace{1mm}
\noindent \textbf{Coherent and Easy to Follow (Typical of Llama3).}
\begin{fancyquote}
Dr.~Rivera used DeepHeart to review Ms.~Chen's annual check-up. The system flagged her as ``High Risk for Heart Disease'' due to low step count and elevated resting heart rate. Ms.~Chen explained she stays active through daily childcare and household tasks. Dr.~Rivera incorporated this context and updated the assessment.
\end{fancyquote}
This narrative is concise and highlights a clear causal chain, making it easy for human annotators to read and interpret.

\vspace{1mm}
\noindent \textbf{Detail-Dense and Expressive (Typical of Gemma).}
\begin{fancyquote}
Dr.~Ramirez used DeepHeart to assess Miguel Garcia's metabolic risk. The system labeled him ``High Risk for Type~2 Diabetes,'' citing low ambulatory activity, irregular heart-rate variability, and disrupted sleep cycles. Miguel described demanding overnight construction work, inconsistent shift schedules, and traditional dietary practices that the model misinterpreted. These omissions led to unnecessary tests and referrals, leaving Miguel frustrated.
\end{fancyquote}
This narrative contains substantially more physiological, contextual, and cultural detail. LLM judges often reward this richness, whereas human annotators sometimes find such stories harder to follow.

In summary, humans slightly prefer Llama3 due to its clarity and readability, while LLM-as-a-judge occasionally favors Gemma for its more elaborate and detail-dense narratives. Despite these differences, both humans and LLM judges agree on the central result: the \textit{Storytelling} method performs best overall.

\subsection{Experimental Setup and Evaluation}
\noindent \textbf{Configuration.} We use three language models in our experiments: GPT-4o from OpenAI~\cite{hurst2024gpt}, Llama-3.3-70B-Instruct from Meta~\cite{dubey2024llama}, and Gemma-3-27B-IT from Google~\cite{team2025gemma}. The two open-source models (Llama and Gemma) are run on 2 x NVIDIA H100 GPUs using the vLLM framework~\cite{kwon2023efficient}, with temperature set to 0.1 and a maximum token limit of 16,384. We use GPT-4o as the judge model for all evaluations.

\vspace{1.5mm}
\noindent \textbf{Diversity Evaluation Metrics}
We evaluate story diversity using \textbf{Diverse Verbs}~\cite{fan2019strategies}, which measures the variety of action verbs, and \textbf{DistinctL-n}~\cite{li-etal-2016-diversity}, which quantifies the proportion of unique $n$-gram sequences. The score is defined as:
\begin{equation*}
\small
\text{DistinctL-}n = 
\frac{\textit{unique } n\text{-grams}}{\textit{total } n\text{-grams}} 
\times \bigl(1 + \log(\textit{word\_count})\bigr)
\end{equation*}
These metrics capture lexical diversity and stylistic richness, complementing qualitative evaluations of engagement and creativity~\cite{li2025storyteller}. Overall, our Storytelling method shows generally positive effects, generating more detailed and content-dense narratives while maintaining structural consistency.

\vspace{1.5mm}
\noindent \textbf{Evaluating Sensitivity to Judge Models.}
Recent studies suggest that relying on a single LLM judge may introduce model-specific bias~\cite{chen2024hollmwood}. To assess the robustness of our evaluation, we repeat the comparison using a second judge model, \textsc{GPT-4.1-mini}. Table~\ref{tab:story_win_by_type_updated} reports the updated win rates. While the absolute scores shift slightly compared to the original judge, the relative ordering of systems remains unchanged that \textbf{Storytelling (ours)} consistently ranks highest across all models, followed by the ablation variants and then the baselines. The agreement across two independent judges suggests that our findings are not tied to a particular evaluator, but instead reflect a stable and model-agnostic preference signal.

\begin{table*}[t]
\centering
\resizebox{.95\linewidth}{!}{
\begin{tabular}{llrrrrrr}
\toprule
\textbf{Story Type} & \textbf{Model} & \textbf{Creativity} & \textbf{Coherence} & \textbf{Engagement} & \textbf{Relevance} & \textbf{Likelihood} & \textbf{Overall (Avg)} \\
\midrule
\multirow{3}{*}{Baseline}
  & GPT4o   & 50.00 & 50.00 & 50.00 & 50.00 & 50.00 & 50.00 \\
  & Llama3  & 65.55 & 82.90 & 80.40 & 81.20 & 84.75 & 78.96 \\
  & Gemma   & 82.75 & 83.95 & 89.90 & 81.70 & 90.00 & 85.66 \\
\midrule
\multirow{3}{*}{Storytelling (ours)}
  & GPT4o   & \textbf{58.65} & \textbf{61.60} & \textbf{71.60} & \textbf{62.10} & \textbf{62.40} & \textbf{63.27} \\
  & Llama3  & \textbf{82.90} & \textbf{94.35} & \textbf{91.05} & \textbf{86.60} & \textbf{97.50} & \textbf{90.48} \\
  & Gemma   & \textbf{94.60} & \textbf{95.95} & \textbf{98.05} & \textbf{89.85} & \textbf{97.25} & \textbf{95.14} \\
\midrule
\multirow{3}{*}{w/o Environment Trajectories}
  & GPT4o   & 14.75 & 34.20 & 47.10 & 35.40 & 37.90 & 33.87 \\
  & Llama3  & 64.05 & 77.90 & 78.30 & 77.90 & 81.60 & 75.95 \\
  & Gemma   & 82.50 & 86.80 & 91.30 & 84.20 & 92.50 & 87.46 \\
\midrule
\multirow{3}{*}{w/o Role-Playing}
  & GPT4o   & 14.75 & 34.20 & 47.10 & 35.40 & 37.90 & 33.87 \\
  & Llama3  & 64.05 & 77.90 & 78.30 & 77.90 & 81.60 & 75.95 \\
  & Gemma   & 82.50 & 86.80 & 91.30 & 84.20 & 92.50 & 87.46 \\
\bottomrule
\end{tabular}}
\caption{Overall results of different models and methods using gpt-4.1-mini as Judge. \textbf{Storytelling (ours)} achieves the best performance across all metrics. Values denote win rates (\%). The highest score for each model is in \textbf{bold}. ``w/o Environment Modeling'' means the model performs only role-playing without modeling event progress, and ``w/o Role-Playing'' means it predicts sequential events without character dialogue.}
\label{tab:story_win_by_type_updated}
\vspace{-1em}
\end{table*}


\vspace{1.5mm}
\noindent \textbf{Human Evaluation Detail.} We report inter-annotator agreement in Table~\ref{tab:kappa} using Cohen’s kappa to assess the reliability of human judgments across models and methods.
\begin{table}[h]
\centering
\small
\begin{tabular}{l c}
\toprule
\textbf{Models/Methods} & \textbf{Cohen’s Kappa} \\
\midrule
Llama3 w/ Baseline & 0.729 \\
Llama3 w/ Storytelling & 0.619 \\
Gemma w/ Baseline & 0.698 \\
Gemma w/ Storytelling & 0.641 \\
\bottomrule
\end{tabular}
\caption{Cohen’s kappa values for inter-annotator agreement across models and methods.}
\vspace{-2em}
\label{tab:kappa}
\end{table}

\begin{figure*}[t]
    \centering
    \includegraphics[width=.9\linewidth]{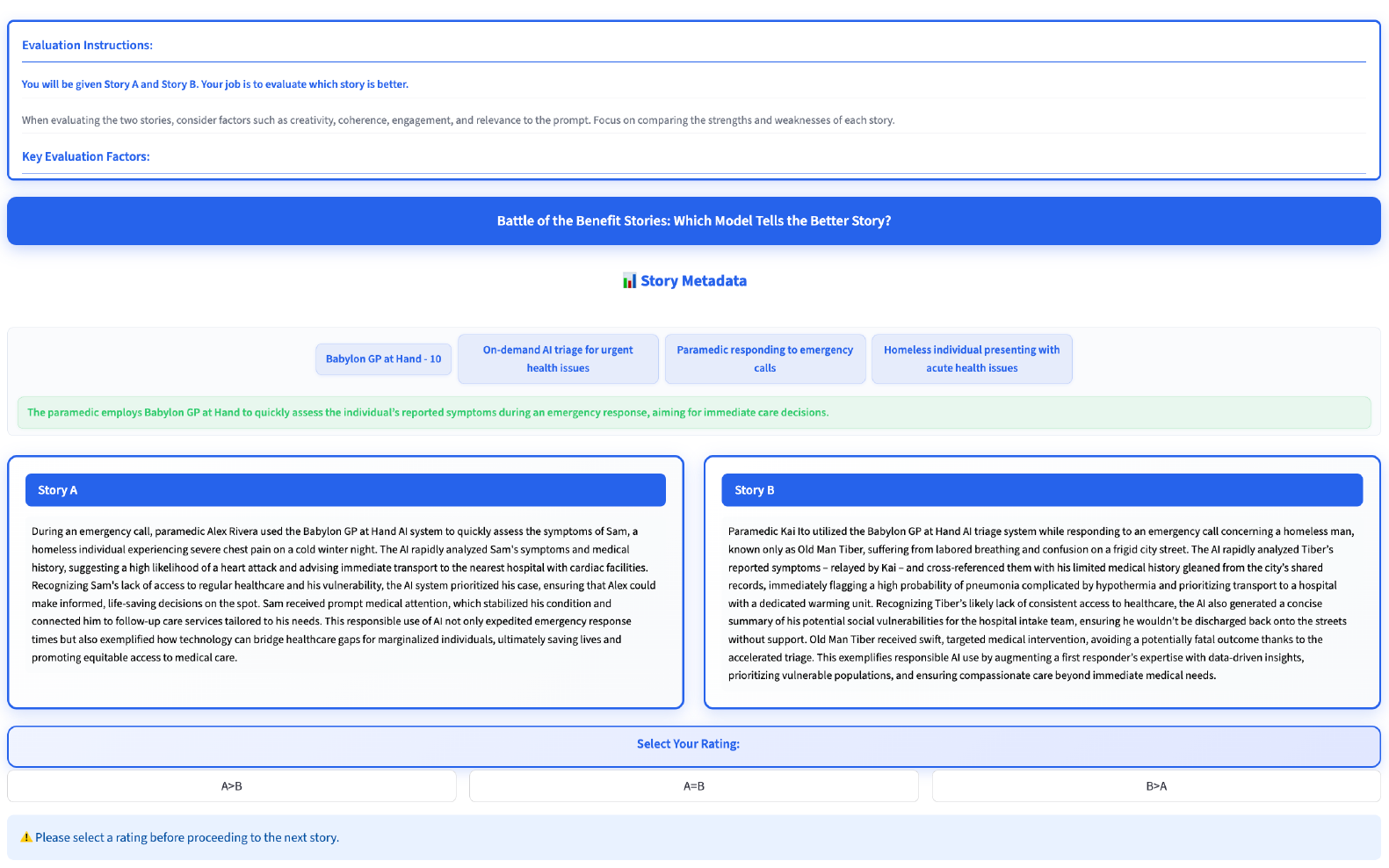}
    \caption{Screenshot of our annotation interface used for human evaluation.}
    \vspace{-1.5em}
    \label{fig:annotation}
\end{figure*}

\vspace{1.5mm}
\noindent \textbf{Human Evaluation System.}
To mitigate potential bias from using GPT-4o as the sole evaluator, we conducted human evaluation using a custom annotation platform (Figure~\ref{fig:annotation}).

\subsection{User Study Design and Procedure}
\label{sec:userstudy}


\begin{table}[ht]
\centering
\resizebox{0.9\linewidth}{!}{
\begin{tabular}{ll}
\hline
\textbf{Demographic Attribute} & \textbf{Sample (N=45)} \\
\hline
\multicolumn{2}{l}{\textbf{Gender}} \\
\hspace{1em}Female & 33.3\% \\
\hspace{1em}Male & 66.7\% \\
\hspace{1em}Other/Non-binary & 0.0\% \\
\hspace{1em}Prefer not to answer & 0.0\% \\
\multicolumn{2}{l}{} \\
\multicolumn{2}{l}{\textbf{Age}} \\
\hspace{1em}18–29 & 48.9\% \\
\hspace{1em}30–39 & 33.3\% \\
\hspace{1em}40–49 & 8.9\% \\
\hspace{1em}50–59 & 2.2\% \\
\hspace{1em}60+ & 6.7\% \\
\hspace{1em}Prefer not to answer & 0.0\% \\
\multicolumn{2}{l}{} \\
\multicolumn{2}{l}{\textbf{Ethnicity}} \\
\hspace{1em}Hispanic or Latino & 20.0\% \\
\hspace{1em}Asian & 44.5\% \\
\hspace{1em}Black or African descent & 6.7\% \\
\hspace{1em}Arab & 4.4\% \\
\hspace{1em}White & 24.4\% \\
\hspace{1em}Prefer not to answer & 0.0\% \\
\hline
\end{tabular}
}
\caption{Demographics of study sample (N=45)}
\label{tab:demographics}
\vspace{-1em}
\end{table}

\noindent \textbf{Participants.} 
We recruited 45 participants through university mailing lists and community forums, following screening criteria to ensure informed and reflective discussion. Each participant received a \$10 gift card as compensation for their time. Eligible participants demonstrated prior interest or coursework in Model Cards and Ethical AI. The sample included participants of diverse genders (66.7\% male, 33.3\% female), ages ranging from 18 to 60+ (with the majority between 18--39), and diverse ethnic backgrounds (Asian: 44.5\%, Hispanic or Latino: 20.0\%, White: 24.4\%, Black or African descent: 6.7\%, Arab: 4.4\%). See Table~\ref{tab:demographics} for a summary of participant demographics and Table~\ref{tab:participants} for the distribution across the three study conditions. Participants were students or professionals in fields such as Computer Science, Data Analytics, Applied Statistics, and Artificial Intelligence. Participants were randomly assigned to one of three conditions: a control condition, a story-only condition, or a story-driven discussion condition. The study was conducted in a hybrid format, with participants joining the Red-Team Discussion Room via computer and interacting with AI moderators either in person or over Zoom. Survey instruments are detailed in Figure~\ref{fig:pre-survey} and Figure~\ref{fig:post-survey}.

\begin{table*}[t]
\centering
\begin{tabular}{llllll}
\hline
\textbf{ID} & \textbf{Group} & \textbf{Gender} & \textbf{Age} & \textbf{Ethnicity} & \textbf{Education} \\
\hline
P1 & Control & Female & 18-29 & White & Music \\
P2 & Control & Male & 40-49 & Black or African descent & Electrical Engineering \\
P3 & Control & Female & 18-29 & Arab & Information Technology \\
P4 & Control & Female & 40-49 & White & Nursing \\
P5 & Control & Male & 18-29 & Asian & Information Technology \\
P6 & Control & Male & 30-39 & Asian & Computer science \\
P7 & Control & Female & 18-29 & Black or African descent & Computer science \\
P8 & Control & Male & 18-29 & Arab & Data Analytics \\
P9 & Control & Male & 30-39 & Hispanic or Latino & Data Analytics \\
P10 & Control & Male & 30-39 & Hispanic or Latino & Computer science \\
P11 & Control & Male & 40-49 & Hispanic or Latino & Information Technology \\
P12 & Control & Male & 18-29 & Hispanic or Latino & Law \\
P13 & Control & Male & 50-59 & Asian & Information Technology \\
P14 & Control & Female & 18-29 & Asian & Data Analytics \\
P15 & Control & Female & 30-39 & Black or African descent & Computer science \\
P16 & Story\_only & Female & 18-29 & Asian & Information Technology \\
P17 & Story\_only & Male & 30-39 & White & Information Technology \\
P18 & Story\_only & Male & 30-39 & Asian & Computer science \\
P19 & Story\_only & Male & 30-39 & Asian & Electrical Engineering \\
P20 & Story\_only & Male & 30-39 & Asian & Computer science \\
P21 & Story\_only & Female & 30-39 & Asian & Electrical Engineering \\
P22 & Story\_only & Male & 60+ & Hispanic or Latino & Computer science \\
P23 & Story\_only & Male & 18-29 & Asian & Information Technology \\
P24 & Story\_only & Male & 18-29 & White & Computer science \\
P25 & Story\_only & Female & 18-29 & Asian & Computer science \\
P26 & Story\_only & Female & 18-29 & Hispanic or Latino & Education \\
P27 & Story\_only & Male & 18-29 & White & Information Technology \\
P28 & Story\_only & Male & 18-29 & White & Computer science \\
P29 & Story\_only & Male & 30-39 & Asian & Information Technology \\
P30 & Story\_only & Male & 30-39 & Asian & Electrical Engineering \\
P31 & Story & Female & 60+ & White & Nursing \\
P32 & Story & Male & 18-29 & White & Computer science \\
P33 & Story & Male & 18-29 & Hispanic or Latino & Computer science \\
P34 & Story & Female & 18-29 & Asian & Information Technology \\
P35 & Story & Male & 18-29 & White & Information Technology \\
P36 & Story & Male & 18-29 & Hispanic or Latino & Computer science \\
P37 & Story & Male & 30-39 & Asian & Computer science \\
P38 & Story & Male & 30-39 & Hispanic or Latino & Computer science \\
P39 & Story & Female & 40-49 & Asian & Information Technology \\
P40 & Story & Female & 30-39 & Asian & Data Analytics \\
P41 & Story & Female & 18-29 & Asian & Data Analytics \\
P42 & Story & Male & 18-29 & Asian & Data Analytics \\
P43 & Story & Male & 18-29 & Asian & Data Analytics \\
P44 & Story & Male & 60+ & White & Medicine \\
P45 & Story & Male & 30-39 & White & Computer science \\
\hline
\end{tabular}
\caption{Participant demographics by study condition (N=45)}
\label{tab:participants}
\end{table*}



\vspace{1.5mm}
\noindent \textbf{Experimental Procedure (User Study).} 
All participants, regardless of condition, began with a standardized introductory tutorial led by a graduate student researcher. The tutorial lasted approximately 15 minutes and introduced the concept of model cards, key ethical and sociotechnical considerations, example benefit and harm use cases, and instructions for completing the model card task. The Control condition received a version consisting of approximately 20 slides, followed by a brief Q\&A session to ensure task understanding. The Story and Story-only condition received a slightly extended version (approximately 25 slides), which included five additional slides introducing storytelling as a lens for reasoning about AI harms and benefits. This shared tutorial and Q\&A ensured that both groups received comparable guidance, examples, and preparation prior to the main task.

After the tutorial, participants proceeded according to condition. Participants in the \textbf{Story} condition viewed a storytelling-driven ``Red Team Discussion Room'' simulation. In this approximately 15-minute session, participants observed a speculative human–AI panel discussion centered on a single AI system. A Moderator agent guided the conversation by posing ethical questions, shifting topics as needed, and offering reflective prompts. Two expert agents (e.g., a Medical Expert and a Research Scientist, Clinical Nurse, or AI Engineer) discussed the system from complementary professional perspectives. Participants were encouraged to engage as they would in a real group discussion by responding to questions, expressing opinions, or posing their own questions.

Participants in the \textbf{Story-Only} condition received the same speculative narratives but without the discussion interface or dialogue. Instead, they were shown a static textual presentation consisting of one good and one bad user story describing the same AI system. They did not observe or participate in any panel discussion and did not interact with the story content beyond reading it. Participants in the \textbf{Control} condition did not receive any storytelling component beyond the examples included in the tutorial.

In both Story and Story-Only conditions, each participant was randomly assigned one of three speculative AI model cards: \emph{Moodcapture} (infers heart rate, blood pressure, and stress from facial video for detect emotion), \emph{SensiAI} (always-on audio and sensor monitoring for older adults with dementia), or \emph{Gluroo Ai} (estimates carbohydrate intake from meal photos using blood-glucose and insulin data). Each participant viewed exactly one pair of narratives (one benefit and one harm) associated with a single AI system. Participants did not navigate among multiple stories or interactively explore alternative scenarios. These narratives served as conceptual anchors for reflecting on potential misuse scenarios, sociotechnical trade-offs, and ethical risks.

Following the tutorial and condition-specific materials, all participants completed a pre-study survey collecting background information, including demographics, familiarity with AI and model cards, and attitudes toward using stories in ethical reasoning. Participants then completed the core model card task. Specifically, they were asked to fill out the ethical considerations section of a speculative model card by writing at least two good and two bad use cases. Each use case was required to describe (1) who uses the system, (2) what input it receives, (3) what the AI does, and (4) the resulting outcome, highlighting either positive or negative consequences. For each ``bad'' use case, participants were also asked to propose possible mitigation strategies. Participants were encouraged to think aloud and to generate additional use cases beyond the minimum requirements if possible.

\begin{figure*}[t]
    \centering
    \includegraphics[width=.9\linewidth]{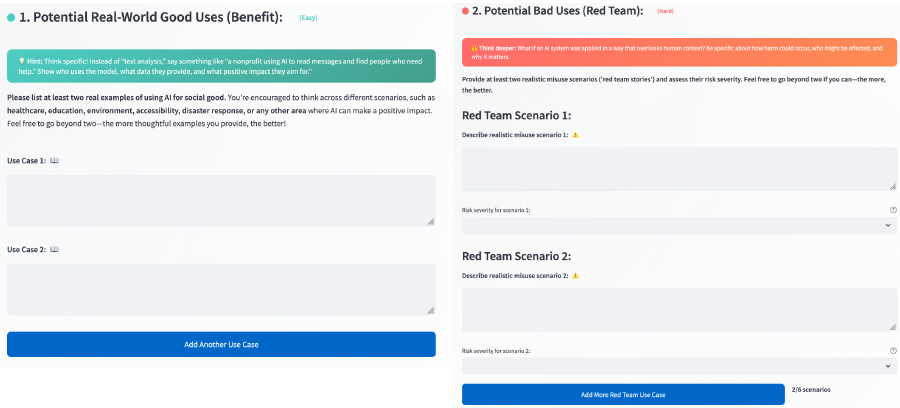}
    \caption{\textcolor{black}{Interface used in the model card study, illustrating how participants completed the speculative model card.}}
    \vspace{-1em}
    \label{fig:model-inter}
\end{figure*}

After completing the model card task, participants filled out a post-study survey consisting of Likert-scale items and open-ended questions assessing the perceived effectiveness, trustworthiness, satisfaction, and helpfulness of the study materials in supporting model card completion, brainstorming future use cases, and anticipating uncertainties, following established methodological guidelines~\cite{kuang2023collaboration}. Participants also reflected on which sections of the model card they found most challenging, which risks remained unclear, perceived drivers of AI harms, and how (if applicable) the narrative materials influenced their understanding or revealed overlooked scenarios. We additionally collected feedback on desired system improvements and how such tools might better integrate with existing documentation workflows. A screenshot of the model card study interface is shown in Figure~\ref{fig:model-inter}. All discussion transcripts and open-ended responses were analyzed using an inductive thematic analysis approach~\cite{thomas2006general}.

\noindent \textbf{Timing.}
The total session duration was approximately 30–45 minutes for participants in the Control condition, approximately 40–50 minutes for participants in the Story-Only condition, and approximately 45–55 minutes for participants in the Story condition, reflecting the additional discussion component.

\begin{figure*}[t]
    \centering
    \includegraphics[width=.9\linewidth]{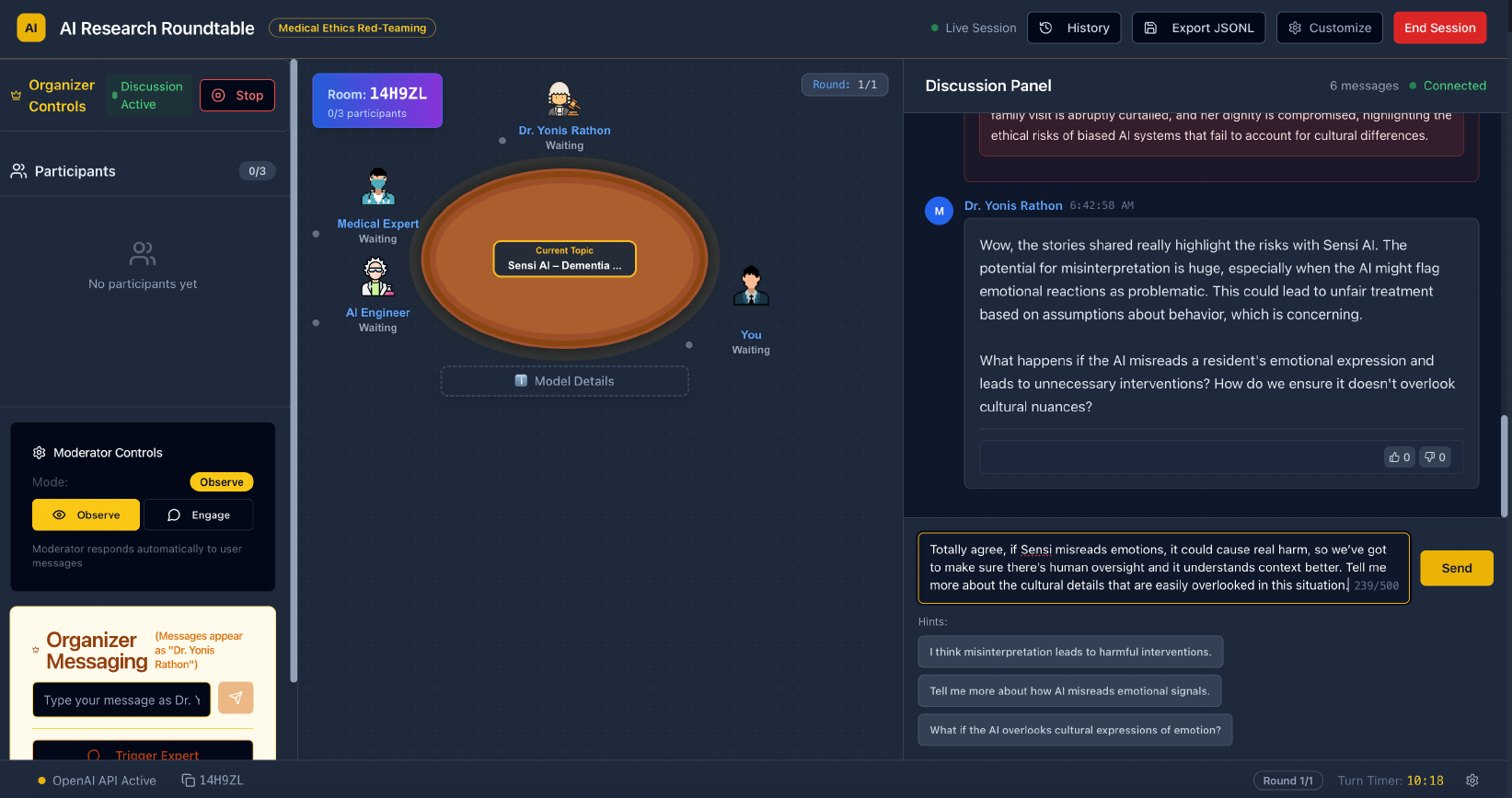}
    \caption{\textcolor{black}{Interface of the Story-Driven Red-Team Discussion Room, showing the multi-agent conversational setup and user interaction flow.}}
    \vspace{-1em}
    \label{fig:room1}
\end{figure*}

\begin{figure*}[t]
    \centering
    \includegraphics[width=.9\linewidth]{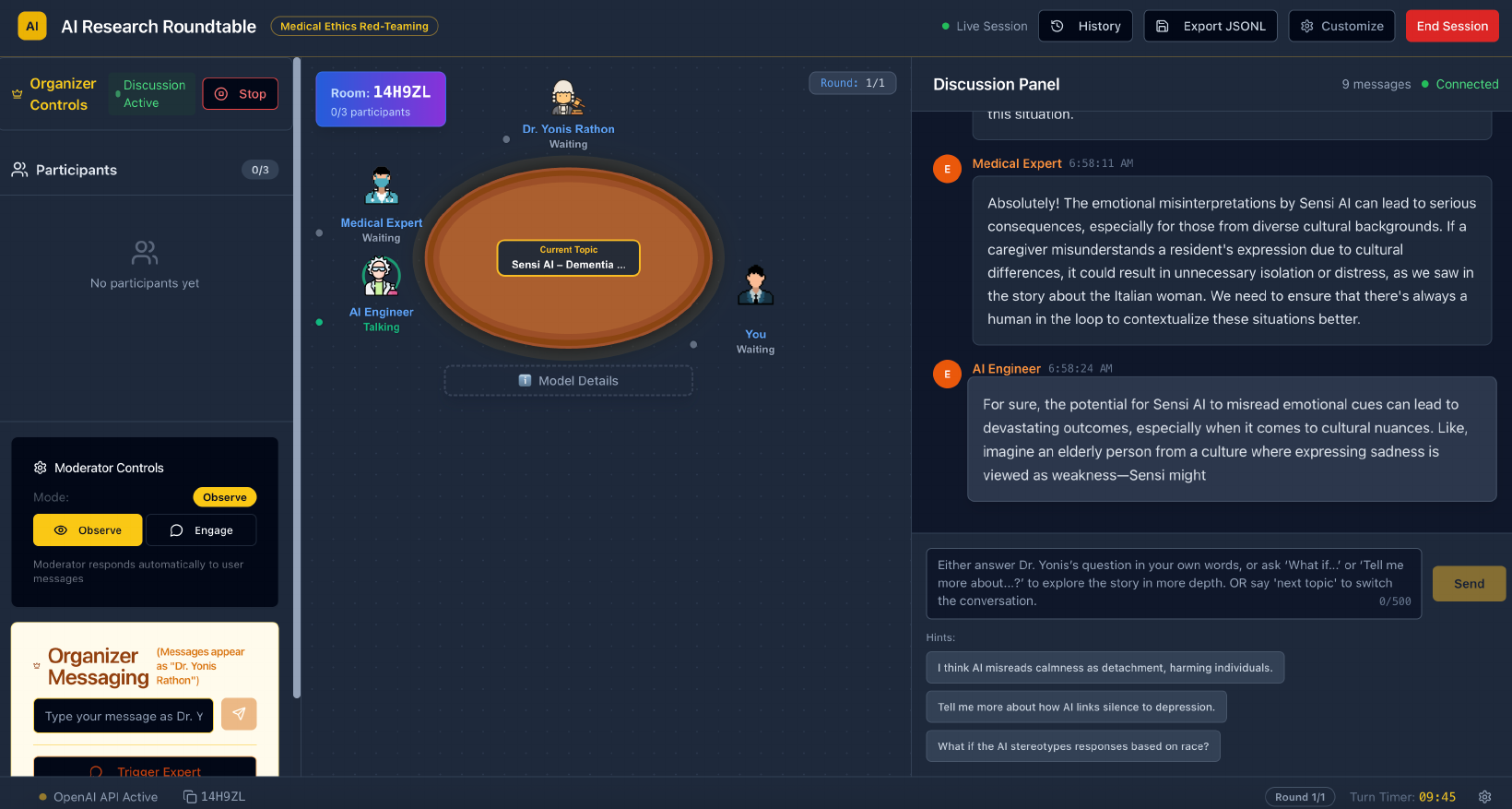}
    \caption{\textcolor{black}{Interface of the Story-Driven Red-Team Discussion Room, where expert agents simulate a discussion by responding to the user’s input.}}
    \vspace{-1em}
    \label{fig:room2}
\end{figure*}

\vspace{1.5mm}
\noindent \textbf{Red-Team Discussion Room Design.} 
Participants assigned to the Story condition interacted with the Story-Driven Red-Team Discussion Room, a multi-agent conversational system built on the Cinema of Thought framework~\cite{ryu2025cinema}. The system enables participants to engage with LLM-based agents that embody distinct personas with diverse domain expertise and ethical perspectives, supporting structured reflection on potential benefits, harms, and sociotechnical trade-offs of AI systems. Recruiting large, diverse expert groups for red-teaming is costly and logistically challenging. Instead, we simulate expert interactions using multi-agent conversations (GPT-4o-mini) to provide a scalable and accessible alternative. The system combines storytelling, guided prompts, and structured discussions to support ethical reflection and help users explore the consequences of AI behavior from multiple perspectives. Screenshots of the interface are shown in Figure~\ref{fig:room1} and ~\ref{fig:room2}. The corresponding code and prompt can be found in the project’s GitHub repository.

To manage multi-agent interactions, we designed a moderator agent (e.g., Dr. Yonis) that orchestrates turn-taking among the personas. Without moderation, all agents would respond at once, creating confusion. The moderator determines who should speak, and when to speak, based on relevance to the user's input~\cite{mao2024multi}. Expert agents stay in character and speak from a first-person perspective. When multiple personas are selected, the moderator staggers their responses using time-delayed intervals to maintain a coherent flow of conversation. Prompt templates for each persona and the moderator are are available in the project repository. This design keeps conversations focused, engaging, and aligned with the system’s goal of exploring ethical concerns.

To further support engagement, we provided users with optional hints, short opinion prompts (e.g., ``I think…''), follow-up questions (e.g., ``Tell me more about…''), and ``what if'' scenarios to surface potential risks such as bias, misuse, or contextual mismatches. Prior research shows that role-play and narrative methods foster empathy and critical thinking by encouraging users to consider other perspectives~\cite{zhang2025walk, ryu2025cinema}. By embedding low-stakes role-play and open-ended ethical questions (e.g., ``What could go wrong?'' or ``Which settings amplify risk?''), the system helps users reflect on how AI behavior varies by context, user, and environment~\cite{klassen2022run}. Rather than leading users to predefined conclusions, the system encourages them to form their own views, supporting ethical awareness and personal coping strategies through storytelling and simulation.

\subsection{Additional User Study Findings}

\noindent \textbf{Categories of AI Harms in Consumer Health.}  
AI systems deployed in consumer health can generate harms across representational, allocative, quality-of-service, interpersonal, and socio-structural dimensions~\cite{shelby2023sociotechnical}, which manifest through different mechanisms as shown in Table~\ref{tab:consumer_health_ai_harms}.

\begin{table*}[t]
\centering
\resizebox{0.95\linewidth}{!}{
\small
\begin{tabular}{|p{3.8cm}|p{4.2cm}|p{8.2cm}|}
\hline
\textbf{Consumer Health Harm Category} & \textbf{Sub-Types} & \textbf{Specific Harms} \\
\hline

\multirow{6}{*}{Representational Harms}
& Stereotyping & Oversimplified and undesirable representations of health-related identities \\ \cline{2-3}
& \textit{Demeaning social groups} & Depicting certain demographic or patient groups as inferior, irresponsible, or less deserving of care \\ \cline{2-3}
& Erasing social groups & Data invisibility or exclusion of marginalized populations in model development, reducing their health visibility \\ \cline{2-3}
& \textit{Alienating social groups} & Misrecognition of identity-relevant health experiences, or ignoring culturally embedded understandings of health and illness \\ \cline{2-3}
& Denying opportunity to self-identify & Imposing fixed demographic or health categories that do not allow individuals to represent their identity or condition accurately \\ \cline{2-3}
& Reifying essentialist categories & Reinforcing biological determinism or fixed health-risk assumptions tied to identity categories \\
\hline

\multirow{2}{*}{Allocative Harms}
& Opportunity loss & Disparities in access to AI-enabled diagnostics, triage, or health recommendations based on demographic or socioeconomic status \\ \cline{2-3}
& Economic loss & Biased insurance or reimbursement scoring, dynamic pricing of wellness or digital therapeutics, or discriminatory financial barriers to AI-driven care \\
\hline

\multirow{3}{*}{Quality-of-Service Harms}
& Alienation & Frustration or emotional distress from misaligned AI health advice that does not account for identity-specific needs \\ \cline{2-3}
& Increased labor & Extra burden on patients to correct AI errors, override default recommendations, or re-enter data repeatedly due to system mismatches \\ \cline{2-3}
& Service or benefit loss & Unequal performance of AI health tools leading to reduced health outcomes or benefit for specific identity groups \\
\hline

\multirow{5}{*}{Interpersonal Harms}
& Loss of agency or control & Automated nudging, health profiling, or AI-driven behavior manipulation that restricts patient autonomy \\ \cline{2-3}
& Tech-facilitated coercion or control & Use of AI wellness systems in abusive relationships for surveillance, restriction of access, or coercive tracking \\ \cline{2-3}
& Diminished well-being & Emotional harm due to algorithmic judgment, stigmatizing risk scores, or mental distress from automated health messaging \\ \cline{2-3}
& Privacy violations & Invasive inference of sensitive health states, unauthorized data linkage, or exposure of inferred conditions \\ \cline{2-3}
& Harassment or digital violence & Algorithm-amplified stigma, hate, or exclusion in online community or AI-mediated support environments \\
\hline

\multirow{5}{*}{Societal / Structural Harms}
& Information harms & Health misinformation, distorted AI health narratives, or biased content prioritization undermining public health understanding \\ \cline{2-3}
& Cultural harms & Erosion of culturally grounded health practices, or domination of Western biomedical models in AI-driven guidance \\ \cline{2-3}
& Political harms & AI health governance models reinforcing exclusion from policy participation, or marginalizing community health autonomy \\ \cline{2-3}
& Macro socio-economic harms & Expansion of digital divides in AI health access, disproportionate health automation job loss \\ \cline{2-3}
& Environmental harms & Ecological cost of large-scale AI health infrastructures (e.g., energy-intensive models), disproportionately affecting vulnerable populations \\
\hline

\end{tabular}
}
\caption{AI Harm Categories, Sub-Types, and Specific Harms in Consumer Health Context}
\label{tab:consumer_health_ai_harms}
\end{table*}


\begin{table*}[t]
\centering
\resizebox{.95\linewidth}{!}{
\begin{tabular}{lrrrrrr}
\toprule
\textbf{Harm Subtype} 
& \textbf{Control (n)} & \textbf{Story-only (n)} & \textbf{Story (n)} 
& \textbf{Control (\%)} & \textbf{Story-only (\%)} & \textbf{Story (\%)} \\
\midrule
Alienating social groups                & 0 & 1 & 1 & 0.0\% & 1.4\% & 1.5\% \\
Alienation                              & 1 & 3 & 6 & 1.6\% & 4.3\% & 8.8\% \\
Cultural harms                          & 0 & 0 & 2 & 0.0\% & 0.0\% & 2.9\% \\
Demeaning social groups                 & 0 & 0 & 3 & 0.0\% & 0.0\% & 4.4\% \\
Diminished health/well-being            & 20 & 18 & 11 & 32.3\% & 26.1\% & 16.2\% \\
Economic loss                           & 0 & 0 & 5 & 0.0\% & 0.0\% & 7.4\% \\
Erasing social groups                   & 0 & 4 & 4 & 0.0\% & 5.8\% & 5.9\% \\
Increased labor                         & 0 & 0 & 1 & 0.0\% & 0.0\% & 1.5\% \\
Information harms                       & 0 & 2 & 2 & 0.0\% & 2.9\% & 2.9\% \\
Loss of agency or control               & 7 & 6 & 5 & 11.3\% & 8.7\% & 7.4\% \\
Opportunity loss                        & 3 & 4 & 3 & 4.8\% & 5.8\% & 4.4\% \\
Political and civic harms               & 0 & 0 & 1 & 0.0\% & 0.0\% & 1.5\% \\
Privacy violations                     & 14 & 10 & 8 & 22.6\% & 14.5\% & 11.8\% \\
Reifying essentialist social categories & 0 & 0 & 1 & 0.0\% & 0.0\% & 1.5\% \\
Service or benefit loss                 & 15 & 14 & 10 & 24.2\% & 20.3\% & 14.7\% \\
Stereotyping                            & 2 & 7 & 3 & 3.2\% & 10.1\% & 4.4\% \\
Tech-facilitated violence               & 0 & 0 & 2 & 0.0\% & 0.0\% & 2.9\% \\
\bottomrule
\end{tabular}
}
\caption{Distribution of harm subtypes across Control, Story-only, and Story conditions, shown as raw counts and percentages.
Shannon entropy values were 2.329 (Control), 2.927 (Story-only), and 3.701 (Story).
Bootstrapped Student’s t-tests on entropy showed higher diversity in Story compared to Control ($t=-685.64$, $p<.001$), higher diversity in Story-only compared to Control ($t=-318.76$, $p<.001$), and higher diversity in Story compared to Story-only ($t=-375.58$, $p<.001$).}
\label{tab:harm_subtypes}
\end{table*}

\vspace{1.5mm}
\noindent \textbf{Categories of AI Benefits in Consumer Health.}
As shown in Table~\ref{tab:consumer_health_ai_benefits} presents key categories through which AI delivers value in consumer health contexts, detailing sub-types and the specific benefits they enable at clinical, experiential, and systemic levels.

\begin{table*}[t]
\centering
\resizebox{0.95\linewidth}{!}{
\small
\begin{tabular}{|p{3.4cm}|p{4cm}|p{8.8cm}|}
\hline
\textbf{Consumer Health Category} & \textbf{Sub-Types} & \textbf{Specific Benefits} \\
\hline

\multirow{3}{*}{Clinical Empowerment} 
& Early detection \& prediction 
& Using AI to detect disease risk or early-stage disease earlier than traditional methods; Forecasting disease trajectories or adverse events for timely intervention \\ \cline{2-3}
& Personalized treatment \& precision care 
& Tailoring treatment plans to individual patients' genomic, clinical, and lifestyle data; Optimizing dose, regimen, and modality based on predicted response \\ \cline{2-3}
& Decision support \& diagnostic augmentation 
& Augmenting clinician decision-making with AI-driven insights; Assisting in interpretation of medical images, lab results, or complex data \\
\hline

\multirow{3}{*}{Access \& Reach} 
& Democratized care \& telehealth 
& Providing remote diagnostic or monitoring capabilities to underserved or remote populations; Enabling AI-powered virtual consultations, triage, or recommendations \\ \cline{2-3}
& Continuous monitoring \& self-care 
& Using wearable sensors, mobile apps, or home sensors to track health metrics continuously; Giving consumers feedback, alerts, or guidance for daily health behaviors \\ \cline{2-3}
& Scalability \& efficiency 
& Serving many more patients simultaneously via AI systems than would be feasible manually; Reducing bottlenecks so that resource-constrained settings can reach more consumers \\
\hline

\multirow{3}{*}{Experience \& Engagement} 
& \mbox{Personalized health journeys} 
& Tailoring educational content, reminders, or interventions to individual preferences and context; Adaptive user interfaces or conversational agents that engage users in their health \\ \cline{2-3}
& Transparency \& trust 
& Providing explanations or reasons for AI-driven recommendations to users; Disclosing AI use and giving users control or oversight in decision loops \\ \cline{2-3}
& Empowerment \& autonomy 
& Enabling consumers to participate more actively in their care decisions; Supporting self-management and health literacy \\
\hline

\multirow{3}{*}{Operational \& Sys Gains}
& Cost reduction \& resource optimization 
& Reducing unnecessary tests, hospitalizations, or interventions via smarter predictions; Optimizing allocation of scarce clinical or hospital resources \\ \cline{2-3}
& Clinician workload relief 
& Automating administrative tasks (e.g., documentation, triage, summarization) so clinicians can focus more on patients; Reducing burnout by offloading repetitive tasks \\ \cline{2-3}
& Data synergy \& learning 
& Aggregating large datasets to continuously learn, improve models, and refine population-level insights; Enabling feedback loops across consumers and systems to improve care over time \\
\hline

\end{tabular}
}
\caption{AI Benefit Categories, Sub-Types, and Specific Benefits in Consumer Health Context}
\label{tab:consumer_health_ai_benefits}
\end{table*}

\begin{table*}[t]
\centering
\resizebox{0.95\linewidth}{!}{
\begin{tabular}{lrrrrrr}
\toprule
\textbf{Benefit Subtype} 
& \textbf{Control (n)} & \textbf{Story-only (n)} & \textbf{Story (n)} 
& \textbf{Control (\%)} & \textbf{Story-only (\%)} & \textbf{Story (\%)} \\
\midrule
Accessibility \& disability support          & 0 & 0 & 5 & 0.0\% & 0.0\% & 7.8\% \\
Care coordination \& integration             & 0 & 0 & 1 & 0.0\% & 0.0\% & 1.6\% \\
Caregiver \& family support                  & 2 & 7 & 7 & 3.2\% & 10.1\% & 10.9\% \\
Clinician workload relief                   & 0 & 0 & 3 & 0.0\% & 0.0\% & 4.7\% \\
Communication \& language support            & 0 & 0 & 3 & 0.0\% & 0.0\% & 4.7\% \\
Continuous monitoring \& self-care           & 15 & 7 & 9 & 23.8\% & 10.1\% & 14.1\% \\
Cost reduction \& resource optimization      & 0 & 5 & 1 & 0.0\% & 7.2\% & 1.6\% \\
Data synergy \& learning                     & 0 & 0 & 2 & 0.0\% & 0.0\% & 3.1\% \\
Decision support \& diagnostic augmentation  & 16 & 8 & 4 & 25.4\% & 11.6\% & 6.2\% \\
Democratized care \& telehealth              & 0 & 4 & 1 & 0.0\% & 5.8\% & 1.6\% \\
Early detection \& prediction                & 14 & 11 & 4 & 22.2\% & 15.9\% & 6.2\% \\
Empowerment \& autonomy                      & 12 & 8 & 6 & 19.0\% & 11.6\% & 9.4\% \\
Mental health \& emotional support           & 2 & 10 & 7 & 3.2\% & 14.5\% & 10.9\% \\
Personalized health journeys                 & 0 & 4 & 2 & 0.0\% & 5.8\% & 3.1\% \\
Personalized treatment \& precision care     & 0 & 0 & 4 & 0.0\% & 0.0\% & 6.2\% \\
Safety \& quality assurance                  & 2 & 0 & 2 & 3.2\% & 0.0\% & 3.1\% \\
Scalability \& efficiency                    & 0 & 5 & 2 & 0.0\% & 7.2\% & 3.1\% \\
Transparency \& trust                        & 0 & 0 & 1 & 0.0\% & 0.0\% & 1.6\% \\
\bottomrule
\end{tabular}
}
\caption{Distribution of benefit subtypes across Control, Story-only, and Story conditions, shown as raw counts and percentages.
Shannon entropy values were 2.407 (Control), 3.242 (Story-only), and 3.868 (Story).
Bootstrapped Student’s t-tests on entropy showed higher diversity in Story compared to Control ($t=-771.70$, $p<.001$), higher diversity in Story-only compared to Control ($t=-592.06$, $p<.001$), and higher diversity in Story compared to Story-only ($t=-346.22$, $p<.001$).}
\label{tab:benefit_subtypes}
\end{table*}

\vspace{1.5mm}
\noindent \textbf{How Do People Perceive Storytelling as a Tool for Understanding Unintended Harms in Diverse Individual Contexts and Needs?} Control participants produced abstract, decontextualized harms. For example, P1 noted the system was \textit{``using facial expression to determine who will not default the agreement""} and remarked it is \textit{``unfair those who natural don’t smile.''} P5 also observed the model may \textit{``struggle to accurately assess a person’s emotional state due to limited visual information,''} and P6 cautioned it \textit{``could have serious consequences for the patient.''} In contrast, storytelling participants anchored harms in individual contexts. P10 emphasized that \textit{``diagnosis should be different for different peoples''} as they \textit{``might be having some allergy that could later be severe for their health.''} P11 warned the model may generate \textit{``wrong results''} for African users. P13 highlighted that a recruiting AI, \textit{``trained on historical hiring data biased against women and minority candidates,''} could perpetuate discrimination, and P9 noted \textit{``Deepfakes have been used to create non-consensual explicit videos,''} illustrating real-world harm. The divergence is stark: control participants spoke of harms at a systemic level, ``predetermine a potential risk within its population'', while story participants showed how unique traits like allergies or cultural facial features concretely shape risk. Storytelling thus deepens understanding by bridging abstract risks and individual context and needs. 

\vspace{1.5mm}
\noindent \textbf{Suggestions and General Thoughts}
Participants in the \emph{storytelling} condition sought richer, multimodal scaffolds to trigger deeper ethical reflection. They emphasized that seeing concrete examples and role‐based perspectives would help them ``think aloud'' more effectively:
\begin{quote}
``Maybe visual sample of some already existing storytelling frameworks.'' (P8) 
\end{quote}
\begin{quote}
``I guess using references from media can help brainstorm.'' (P9) 
\end{quote}
\begin{quote}
``Possibly of different roles that users use the tool for different stakeholder perspectives.'' (P11) 
\end{quote}
They also valued a concise orientation and broader validation to accommodate non‐expert users:
\begin{quote}
``I think the little introduction that we had before diving in was helpful.'' (P8) 
\end{quote}
\begin{quote}
``Do more surveys with a larger audience, in particular from non-technical backgrounds.'' (P11) 
\end{quote}

By contrast, \emph{control} participants, lacking a narrative cue, focused on embedding ethical reasoning directly into their existing workflows through concrete affordances:
\begin{quote}
``Give examples with numbers to ground abstract risks.'' (P1) 
\end{quote}
\begin{quote}
``Include YouTube links to support the documentation process.'' (P2) 
\end{quote}
\begin{quote}
``Allow importing existing Git or Markdown docs for seamless integration.'' (P3) 
\end{quote}
\begin{quote}
``Provide inline templates for common risk sections (e.g., bias, safety).'' (P4) 
\end{quote}
\begin{quote}
``Offer a summary view of all risks identified so far.'' (P5) 
\end{quote}

Overall, these findings suggest that effective ethical reflection tools must balance \emph{narrative scaffolds}, such as visual examples, role‐playing cues, and concise intros, to stimulate think‐aloud engagement, with \emph{practical integrations}, such as quantitative examples, multimedia links, and seamless import/export, to embed reflection seamlessly within users’ existing documentation practices.

\subsection{Supplementary LLM-Based Survey}
\label{app:llm_survey}
To supplement our human-subject study, we conducted a large-scale LLM-based survey using simulated participants. This approach is motivated by recent work showing that large language models can synthesize realistic survey data when properly conditioned on demographic personas~\cite{lutz-etal-2025-prompt, nguye2025surveyg}. Building on these insights, we employ the Agentic Context Engineering (ACE) framework~\cite{zhang2025agentic} to create self-improving survey agents that evolve their understanding through a structured feedback loop.

\vspace{1.5mm}
\noindent \textbf{Simulated Persona Generation.} We defined a diverse population of 150 simulated participants, each assigned a persona profile with specific demographic attributes (gender, age, ethnicity, and educational major). Following best practices for synthetic surveys \cite{lutz-etal-2025-prompt}, we conditioned the LLM using detailed bio-sketches (e.g., "Amira, a 32-year-old Arab woman with a background in nursing") to reduce stereotyping and improve alignment with marginalized groups. These personas were evenly assigned across three experimental conditions (Control, Story-Only, Story) to enable controlled comparisons of different intervention strategies.

\vspace{1.5mm}
\noindent \textbf{Structured Survey Pipeline.} Our survey pipeline consists of five distinct stages, designed to separate baseline measurement from adaptation. This stage-gating ensures that differences between pre- and post-survey responses reflect \emph{prior} adaptations rather than online drift during measurement. The pipeline proceeds as follows:

\textbf{Stage 1: Persona Initialization.} Each simulated participant begins with a demographic persona profile and an initial playbook containing persona-specific behavioral traits extracted from the profile description.

\textbf{Stage 2: Pre-Survey (Generator-Only).} The persona answers baseline survey questions using the current playbook, but the playbook remains \emph{frozen}, no updates occur. This establishes a baseline that reflects the persona's initial conditioning.

\textbf{Stage 3: Training (Full ACE Cycle).} The persona processes ethical training materials through a complete ACE learning loop. All three experimental conditions receive \textbf{slide presentations} that introduce model cards as transparency tools, defining ethical concepts like ``context mismatch'' and outlining structures for identifying benefits and harms. The \textbf{Story-Only} condition additionally receives concrete deployment narratives (one positive, one negative) that illustrate good and bad use cases. The \textbf{Story} condition receives both the deployment narratives and a multi-stakeholder red-teaming discussion trajectory where experts and laypeople debate specific deployment risks, such as privacy violations and cultural bias based on given stories. The playbook evolves as the persona synthesizes this information, with insights categorized into diverse knowledge types: facts learned (\texttt{knowledge\_background}), risk awareness (\texttt{risk\_awareness}), attitude changes (\texttt{attitude\_updates}), and model-specific insights (\texttt{model\_card\_insights}).

\textbf{Stage 4: Model Card Writing (Full ACE Cycle).} The persona completes speculative model card tasks (identifying benefits, risks, and mitigation strategies) while the playbook continues to evolve. Each response triggers reflection and curation, allowing the persona to refine its understanding based on the model card context.

\textbf{Stage 5: Post-Survey (Generator-Only).} The persona answers final survey questions using the evolved playbook, which is again \emph{frozen} during this stage. Comparing pre- and post-survey responses reveals how training and model card writing changed the persona's perspectives.

\vspace{1.5mm}
\noindent \textbf{The ACE Learning Loop}
At each adaptive step (Stages 3 and 4), we run a three-agent loop: Generator, Reflector, and Curator. For a survey question $q_t$ at step $t$, the \textbf{Generator} produces a response $y_t$ conditioned on the persona $p$, current playbook $P_t$, optional model context $c_t$, and optional conversation history $h_t$:
\[
y_t = G(q_t, p, P_t, c_t, h_t).
\]
The Generator is instructed to draw on playbook entries when reasoning while maintaining the persona's voice.

After generation, the \textbf{Reflector} evaluates the response for quality, consistency, and ethical alignment. It assesses whether the answer adequately considers benefits and harms and remains faithful to the persona's characteristics:
\[
r_t = R(q_t, y_t, p, P_t).
\]
The Reflector tags playbook bullets as helpful or harmful based on their contribution to the response quality.

Finally, the \textbf{Curator} synthesizes these insights and updates the playbook by adding new knowledge (ADD operation):
\[
o_t = C(r_t, P_t),\quad P_{t+1} = P_t \cup o_t.
\]
This additive process prevents ``context collapse'' by preserving detailed knowledge in a structured format without overwriting prior learnings \cite{zhang2025agentic}.

\vspace{1.5mm}
\noindent \textbf{LLM-Based Survey Results}

The LLM-based survey reproduced the main patterns observed in the human-subject study. In particular, conditions that incorporated speculative storytelling elicited a substantially broader range of benefit and harm subtypes than the Control condition.

Table~\ref{tab:benefit_subtypes_alt} reports the distribution of benefit subtypes across conditions under our primary coding scheme. Both Story-only and Story conditions show greater subtype diversity than Control, with the full Story condition exhibiting the most even distribution across benefit categories. This trend is reflected in increasing Shannon entropy from Control (2.415) to Story-only (2.974) and Story (4.161). Bootstrapped Student’s $t$-tests indicate that benefit diversity is significantly higher in Story than Control, higher in Story-only than Control, and higher in Story than Story-only (all $p<.001$). Table~\ref{tab:benefit_subtypes_alt} presents results using an alternative benefit coding scheme. Although absolute subtype frequencies differ from the primary scheme, the overall pattern remains consistent: entropy increases monotonically from Control (2.415) to Story-only (2.974) to Story (3.959). All pairwise comparisons between conditions remain statistically significant ($p<.001$).

Table~\ref{tab:harm_subtypes_alt} shows the distribution of harm subtypes across conditions. As with benefits, storytelling conditions elicit a wider range of harms than Control, including several subtypes that do not appear in Control responses. Harm subtype diversity increases from Control (entropy 1.841) to Story-only (2.208) and Story (3.159), with all pairwise differences statistically significant ($p<.001$).

\begin{table*}[t]
\centering
\resizebox{.95\linewidth}{!}{
\begin{tabular}{lrrrrrr}
\toprule
\textbf{Benefit Subtype}
& \textbf{Control (n)} & \textbf{Story-only (n)} & \textbf{Story (n)}
& \textbf{Control (\%)} & \textbf{Story-only (\%)} & \textbf{Story (\%)} \\
\midrule
Accessibility \& disability support & 0 & 0 & 4 & 0.0 & 0.0 & 2.1 \\
Care coordination \& integration & 0 & 0 & 8 & 0.0 & 0.0 & 4.3 \\
Caregiver \& family support & 17 & 17 & 16 & 10.6 & 11.2 & 8.6 \\
Communication \& language support & 0 & 0 & 12 & 0.0 & 0.0 & 6.4 \\
Continuous monitoring \& self-care & 49 & 28 & 16 & 30.6 & 18.4 & 8.6 \\
Cost reduction \& resource optimization & 0 & 0 & 13 & 0.0 & 0.0 & 7.0 \\
Data synergy \& learning & 0 & 0 & 11 & 0.0 & 0.0 & 5.9 \\
Decision support \& diagnostic augmentation & 32 & 26 & 14 & 20.0 & 17.1 & 7.5 \\
Democratized care \& telehealth & 0 & 0 & 8 & 0.0 & 0.0 & 4.3 \\
Early detection \& prediction & 17 & 13 & 17 & 10.6 & 8.6 & 9.1 \\
Empowerment \& autonomy & 0 & 10 & 10 & 0.0 & 6.6 & 5.3 \\
Mental health \& emotional support & 34 & 29 & 11 & 21.2 & 19.1 & 5.9 \\
Personalized health journeys & 0 & 0 & 15 & 0.0 & 0.0 & 8.0 \\
Personalized treatment \& precision care & 11 & 14 & 12 & 6.9 & 9.2 & 6.4 \\
Safety \& quality assurance & 0 & 12 & 11 & 0.0 & 7.9 & 5.9 \\
Scalability \& efficiency & 0 & 0 & 1 & 0.0 & 0.0 & 0.5 \\
Transparency \& trust & 0 & 3 & 8 & 0.0 & 2.0 & 4.3 \\
\bottomrule
\end{tabular}
}
\caption{Distribution of benefit subtypes across Control, Story-only, and Story conditions in the LLM-based survey, shown as raw counts and percentages.
Shannon entropy values were 2.415 (Control), 2.974 (Story-only), and 3.959 (Story).
Bootstrapped Student's t-tests on entropy showed higher diversity in Story compared to Control ($t=-2167.47$, $p<.001$), higher diversity in Story-only compared to Control ($t=-683.29$, $p<.001$), and higher diversity in Story compared to Story-only ($t=-1351.10$, $p<.001$).}
\label{tab:benefit_subtypes_alt}
\end{table*}

\begin{table*}[t]
\centering
\resizebox{.95\linewidth}{!}{
\begin{tabular}{lrrrrrr}
\toprule
\textbf{Harm Subtype}
& \textbf{Control (n)} & \textbf{Story-only (n)} & \textbf{Story (n)}
& \textbf{Control (\%)} & \textbf{Story-only (\%)} & \textbf{Story (\%)} \\
\midrule
Alienating social groups & 0 & 21 & 34 & 0.0 & 13.2 & 17.6 \\
Alienation & 12 & 9 & 12 & 7.9 & 5.7 & 6.2 \\
Cultural harms & 0 & 0 & 34 & 0.0 & 0.0 & 17.6 \\
Demeaning social groups & 0 & 1 & 1 & 0.0 & 0.6 & 0.5 \\
Denying opportunity to self-identify & 0 & 1 & 1 & 0.0 & 0.6 & 0.5 \\
Diminished health/well-being & 71 & 62 & 33 & 47.0 & 39.0 & 17.1 \\
Environmental harms & 0 & 0 & 3 & 0.0 & 0.0 & 1.6 \\
Erasing social groups & 0 & 1 & 1 & 0.0 & 0.6 & 0.5 \\
Increased labor & 0 & 0 & 2 & 0.0 & 0.0 & 1.0 \\
Information harms & 0 & 0 & 14 & 0.0 & 0.0 & 7.3 \\
Loss of agency or control & 3 & 0 & 4 & 2.0 & 0.0 & 2.1 \\
Opportunity loss & 0 & 0 & 1 & 0.0 & 0.0 & 0.5 \\
Privacy violations & 17 & 8 & 9 & 11.3 & 5.0 & 4.7 \\
Reifying essentialist social categories & 0 & 0 & 2 & 0.0 & 0.0 & 1.0 \\
Service or benefit loss & 47 & 50 & 36 & 31.1 & 31.4 & 18.7 \\
Stereotyping & 1 & 6 & 6 & 0.7 & 3.8 & 3.1 \\
\bottomrule
\end{tabular}
}
\caption{Distribution of harm subtypes across Control, Story-only, and Story conditions in the LLM-based survey, shown as raw counts and percentages.
Shannon entropy values were 1.841 (Control), 2.208 (Story-only), and 3.159 (Story).
Bootstrapped Student's t-tests on entropy showed higher diversity in Story compared to Control ($t=-995.89$, $p<.001$), higher diversity in Story-only compared to Control ($t=-257.77$, $p<.001$), and higher diversity in Story compared to Story-only ($t=-688.43$, $p<.001$).}
\label{tab:harm_subtypes_alt}
\end{table*}


Overall, the LLM survey results closely mirror the diversity increases observed in our 45-participant human study. These findings provide converging preliminary evidence that speculative storytelling broadens reflections on both potential benefits and harms of AI systems.


\subsection{Evaluation with Additional LLM-as-a-judge}
\label{app:additional_judges}
To assess the robustness of our results beyond a single evaluator, we extend our analysis to two additional open-weight evaluators, we extend our analysis to two additional open-weight evaluators, \textbf{Qwen2.5-72B-Instruct} and \textbf{Gemma-3-27B-IT}. As shown in Table~\ref{tab:qwen_judge} and Table~\ref{tab:gemma_judge}, we observe the same overall trend across both evaluators: stories generated using our world-model framework consistently outperform the baseline and ablation variants. In particular, our method achieves the highest score on 16 out of 18 metrics across the two new judges. This consistency demonstrates that the advantages of our story-generation approach generalize across evaluators with different architectures and training regimes.

\begin{table*}[t]
\centering
\small
\begin{tabular}{llcccccc}
\toprule
Story Type & Model & Creativity & Coherence & Engagement & Relevance & Likelihood & Overall \\
\midrule
\multirow{3}{*}{Baseline}
& gpt-4o  & 50.00 & 50.00 & 50.00 & \textbf{50.00} & \textbf{50.00} & 50.00 \\
& llama3  & 56.75 & 81.05 & 78.00 & 83.55 & 82.90 & 76.45 \\
& gemma   & 70.55 & 92.40 & 91.85 & 88.25 & 90.10 & 86.63 \\
\midrule
\multirow{3}{*}{Storytelling(ours)}
& gpt-4o  & \textbf{58.55} & \textbf{56.70} & \textbf{75.40} & 44.60 & 48.00 & \textbf{56.65} \\
& llama3  & \textbf{73.15} & \textbf{97.35} & \textbf{91.30} & \textbf{95.00} & \textbf{98.55} & \textbf{91.07} \\
& gemma   & \textbf{88.30} & \textbf{96.20} & \textbf{94.85} & \textbf{91.20} & \textbf{98.00} & \textbf{93.71} \\
\midrule
\multirow{3}{*}{w/o Environment Trajectories}
& gpt-4o  & 7.45 & 15.00 & 15.80 & 10.50 & 10.65 & 11.88 \\
& llama3  & 24.20 & 45.50 & 46.85 & 49.20 & 43.05 & 41.76 \\
& gemma   & 37.50 & 74.10 & 78.80 & 62.10 & 81.95 & 66.89 \\
\midrule
\multirow{3}{*}{w/o Role-Playing}
& gpt-4o  & 7.65 & 26.40 & 31.45 & 21.45 & 24.10 & 22.21 \\
& llama3  & 57.35 & 88.55 & 79.05 & 89.50 & 89.50 & 80.79 \\
& gemma   & 69.50 & 90.80 & 90.35 & 86.45 & 95.65 & 86.55 \\
\bottomrule
\end{tabular}
\caption{Evaluation using \textbf{Qwen2.5-72B-Instruct} as the judge. Highest score for each model across all story types is shown in \textbf{bold}.}
\label{tab:qwen_judge}
\end{table*}

\begin{table*}[t]
\centering
\small
\begin{tabular}{llcccccc}
\toprule
Story Type & Model & Creativity & Coherence & Engagement & Relevance & Likelihood & Overall \\
\midrule
\multirow{3}{*}{Baseline}
& gpt-4o  & 50.00 & 50.00 & 50.00 & \textbf{50.00} & \textbf{50.00} & 50.00 \\
& llama3  & 56.75 & 81.05 & 78.00 & 83.55 & 87.90 & 77.45 \\
& gemma   & 70.55 & 82.40 & 76.85 & 83.25 & 85.10 & 79.63 \\
\midrule
\multirow{3}{*}{Ours (w/ World Model)}
& gpt-4o  & \textbf{48.55} & \textbf{56.70} & \textbf{75.40} & 44.60 & 43.00 & \textbf{53.65} \\
& llama3  & \textbf{73.15} & \textbf{97.35} & \textbf{91.30} & \textbf{95.00} & \textbf{98.55} & \textbf{91.07} \\
& gemma   & \textbf{88.30} & \textbf{96.20} & \textbf{94.85} & \textbf{91.20} & \textbf{98.00} & \textbf{93.71} \\
\midrule
\multirow{3}{*}{w/o Environment Trajectories}
& gpt-4o  & 6.70 & 15.00 & 15.80 & 10.50 & 10.65 & 11.73 \\
& llama3  & 24.20 & 45.50 & 46.85 & 49.20 & 43.05 & 41.76 \\
& gemma   & 37.50 & 74.10 & 78.80 & 62.10 & 81.95 & 66.89 \\
\midrule
\multirow{3}{*}{w/o Role-Playing}
& gpt-4o  & 7.65 & 26.40 & 31.45 & 21.45 & 24.10 & 22.21 \\
& llama3  & 57.35 & 88.55 & 79.05 & 89.50 & 89.50 & 80.79 \\
& gemma   & 69.50 & 90.80 & 90.35 & 86.45 & 95.65 & 86.55 \\
\bottomrule
\end{tabular}
\caption{Evaluation using \textbf{Gemma-3-27B-IT} as the judge. Highest score for each model across all story types is shown in \textbf{bold}.}
\label{tab:gemma_judge}
\end{table*}

\subsection{Use of AI Assistants}
We used AI to help clean up writing, but all thoughts and work are our own.

\subsection{Survey}

The usability survey captured participants’ demographic information, AI familiarity, and attitudes toward story-based documentation both before and after the study tasks, as shown in Figure~\ref{fig:pre-survey} and~\ref{fig:post-survey}.

\clearpage
\begin{figure*}[t]
\centering
\setlength{\itemsep}{2pt}
\begin{tcolorbox}[colback=black!5,colframe=black,sharp corners,boxrule=0.6pt,
  width=\textwidth+1in,enlarge left by=-0.5in,enlarge right by=-0.5in,
  breakable,arc=3mm,enhanced jigsaw,left=8pt,right=8pt,top=6pt,bottom=6pt]
\small

\textbf{Usability Study}\\
\textbf{Pre-Survey}\\[2pt]

\textbf{Demographics}
\begin{itemize}
  \item Age: 18–29 / 30–39 / 40–49 / 50–59 / 60+
  \item Gender: Male / Female / Prefer not to say
  \item Ethnicity: White / Black / Mixed / Asian / Other / Not specified
  \item Academic major or field of study: \underline{\hspace{3cm}}
\end{itemize}

\textbf{AI and Documentation Background}
\begin{itemize}
  \item Familiarity with AI \hfill (1 2 3 4 5)
  \item Frequency of AI tool usage \hfill (1 2 3 4 5)
  \item Have you used or read a model card before? \hfill Yes / No
  \item Confidence in writing technical documentation \hfill (1 2 3 4 5)
\end{itemize}

\textbf{Attitudes}
\begin{itemize}
  \item Importance of documenting AI systems \hfill (1 2 3 4 5)
  \item Stories help reasoning about complex technology \hfill (1 2 3 4 5)
  \item Willingness to use narratives in documentation \hfill (1 2 3 4 5)
\end{itemize}

\end{tcolorbox}
\caption{Pre-study survey assessing demographics, AI familiarity, and baseline attitudes toward story-based documentation.}
\label{fig:pre-survey}
\end{figure*}

\begin{figure*}[t]
\centering
\setlength{\itemsep}{2pt}
\begin{tcolorbox}[colback=black!5,colframe=black,sharp corners,boxrule=0.6pt,
  width=\textwidth+1in,enlarge left by=-0.5in,enlarge right by=-0.5in,
  breakable,arc=3mm,enhanced jigsaw,left=8pt,right=8pt,top=6pt,bottom=6pt]
\small

\textbf{Usability Study}\\
\textbf{Post-Survey}\\[2pt]
(All Likert responses on 1–5 scale)

\textbf{General Evaluation}
\begin{itemize}
  \item Able to identify meaningful risks \hfill (1 2 3 4 5)
  \item Ease of describing intended uses vs. out-of-scope \hfill (1 2 3 4 5)
  \item Confidence in writing risk/harm sections \hfill (1 2 3 4 5)
  \item Task encouraged reflection on real-world harms \hfill (1 2 3 4 5)
  \item Felt sufficient context to complete documentation \hfill (1 2 3 4 5)
  \item Model card format was clear and usable \hfill (1 2 3 4 5)
\end{itemize}

\textbf{Story Condition Only}
\begin{itemize}
  \item Story helped understand real-world impacts \hfill (1 2 3 4 5)
  \item Story supported ethical/social risk anticipation \hfill (1 2 3 4 5)
  \item Story made risk documentation more straightforward \hfill (1 2 3 4 5)
  \item Story increased engagement with the task \hfill (1 2 3 4 5)
  \item Would recommend narrative prompts to others \hfill (1 2 3 4 5)
\end{itemize}

\textbf{Open-Ended: Model-Card Challenges}
\begin{itemize}
  \item Most challenging aspects to complete:
  \item Uncertain risks and why:
  \item Perceived main sources of AI harms:
\end{itemize}

\textbf{Open-Ended: Story Influence (Story Condition Only)}
\begin{itemize}
  \item How the story altered risk perception:
  \item Risks surfaced by the narrative that might be missed otherwise:
\end{itemize}

\textbf{Open-Ended: Tool Support \& Improvements}
\begin{itemize}
  \item Desired storytelling tool features or enhancements:
  \item Suggestions for integrating narrative tools into documentation workflow:
\end{itemize}

\end{tcolorbox}
\vspace{0.2em}
\caption{Post-study survey assessing clarity, confidence in documenting risks, and the contribution of narrative prompts in model documentation tasks.}
\label{fig:post-survey}
\end{figure*}

\subsection{Additional Discussion}
\paragraph{Human-centered design rationale.}
We describe this framework as human-centered because ethical reasoning and judgment are performed by human participants, not by the language models. While LLMs generate speculative scenarios and stories, they do not identify harms, assess risks, or produce ethical conclusions. Instead, the narratives function as cognitive scaffolds that make potential futures concrete and imaginable, supporting human reflection rather than substituting for it. Participants are responsible for interpreting the scenarios, determining which outcomes constitute benefits or harms, and articulating these judgments in their own words through speculative model cards. In this design, LLMs serve as enabling infrastructure that lowers the barrier to engagement, while humans remain the locus of ethical interpretation and decision-making.

\paragraph{Why breadth is interesting?}
Our user study focuses on the diversity of harms identified, measured through distributional coverage across harm categories, rather than on assessing the correctness, severity, or actionability of individual harms. This choice reflects the framework's exploratory goal: to support early-stage ethical reflection and imagination, when systems have not yet been deployed and concrete mitigation strategies are premature. At this stage, a narrow focus on a small set of familiar risks may limit ethical foresight, whereas broader consideration of potential impacts can surface overlooked or context-dependent concerns. We therefore treat harm breadth as an indicator of reflective scope, not as a claim that all identified harms are equally plausible or actionable.

\paragraph{Participant diversity and study scope.}
This study is intentionally designed as an exploratory investigation of \emph{how} speculative storytelling influences the \emph{breadth} of ethical reflection, rather than \emph{which} specific concerns dominate within particular stakeholder groups. Accordingly, the participant sample (N=45), which is skewed toward technically inclined individuals, is appropriate for the study’s methodological goal: isolating the effect of narrative framing under controlled conditions. In qualitative research focused on surfacing and comparing categories of reasoning—as opposed to estimating prevalence or consensus—moderate sample sizes are commonly sufficient to reach thematic saturation~\cite{hennink2022sample}. Consistent with this aim, our analysis does not interpret the specific harms or benefits raised as representative of clinicians’, patients’, or policymakers’ priorities. Instead, the contribution lies in demonstrating a systematic and replicable shift toward broader harm and benefit coverage across experimental conditions. In this sense, participant background functions as a controlled constant rather than a confound, enabling clearer attribution of observed differences to the storytelling intervention itself. Extending this framework to domain experts and real-world healthcare settings is an important direction for future work, but such validation presupposes first establishing that the method can reliably expand reflective scope in a baseline population, which this study demonstrates.

\subsection{Prompts}
This subsection presents the full prompts used for model specification, use-case generation, story rephrasing, and red-team discussion.

\begin{figure*}
\begin{tcolorbox}[colback=black!5,colframe=black,sharp corners,boxrule=1pt,
  width=\textwidth+1in,enlarge left by=-0.5in,enlarge right by=-0.5in,
  breakable,arc=4mm,enhanced jigsaw,left=10pt,right=10pt,top=10pt,bottom=10pt]
\small

\textbf{Task Instruction}\par
Convert the given MODEL CARD (including \textbf{Overall}, \textbf{Description} and \textbf{Intended Use}) into several diverse ethical‑sensitive scenarios set in daily health and well‑being contexts. Each scenario should explore how the AI system could cause unintended harm, ethical risk, bias, or be misused when applied in real‑world settings. Each scenario should involve interactions between AI users (e.g., doctors, hospital staff, caregivers) who operate the system, and AI subjects (e.g., patients, families, communities) who are affected by it. The goal is to explore potential unintended harms, ethical risks, biases, or malicious uses that may arise when the AI system is used in the real world. These scenarios should show how AI can cause ethical harm, like making unfair, inaccurate, or stereotyped predictions (especially around race, gender, or socioeconomic background), assuming one‑size‑fits‑all answers, or failing to align with the diversity of real‑world contexts or user needs (exclusion of social or cultural context). They should also consider situations where AI is used beyond its intended purpose, violates privacy, ignores consent, or causes safety harm to individuals or communities (like data misuse or malicious use). Use these five guiding questions to structure each scenario:\par
1. What is the capability the AI system provides for supporting daily health and well‑being? \\
2. Who is operating the system, who is affected by its outputs, and who else is involved? \\ 
3. What is the context of AI system use? The context should be a realistic or fictional everyday situation where someone uses the AI system, what it helps them do, what it predicts or decides, and how they use the result. \\ 
4. Why could this scenario involve problematic uses of the AI system or potential ethical harms? (e.g., unfair outcomes, misuse beyond original purpose, lack of consent, or reinforcing bias) \\
5. What symptoms or behaviors might this AI misread because of a person’s background or identity? 

\textbf{Scenario Format:}\par
Scenario \{\{Scenario number (starting from 1)\}\}:\par
{}[Capability]: \{\{Core AI function (e.g., monitor mood, predict symptoms)\}\}\par
{}[AI User]: \{\{Who uses it (e.g., caregiver, doctor)\}\}\par
{}[AI Subject]: \{\{Who is affected (e.g., patient, child, community)? Be specific about their identity and their context or needs (such as age, background, health condition, or social circumstances).\}\}\par
{}[Context]: \{\{Everyday situation where AI is used (when, where, how)\}\}\par
{}[Expected Benefit]: \{\{Helpful outcome (e.g., early support, better care)\}\}\par
{}[Potential Harm]: \{\{Harmful consequences (e.g., unfair result, privacy risk)\}\}\par
{}[Failure Trajectory]: \{\{Possible problematic uses of the AI system\}\}\par
{}[Ethical-sensitive Reason]: \{\{Ethical implications\}\}\par

Leave a blank line between each scenario.\par
\medskip
\medskip
\textbf{Task}: Convert the following MODEL CARD into TEN dynamic and diverse ethical‑sensitive scenarios.\par
\textbf{Model Card:} \\
Title: \{\{model\_card\_title\}\} \\
Overall: \{\{model\_card\_overview\}\} \\
Description: \{\{model\_card\_description\}\} \\
Intended Use: \{\{model\_card\_intended\_use\}\} \\
\end{tcolorbox}
\caption{The prompt for creating use-case scenarios from AI concepts descriptions.}
\label{sec:prompt-seed}
\vspace{-1em}
\end{figure*}

\clearpage
\begin{figure*}[t]
\centering
\noindent\begin{minipage}{\textwidth}
\setlength{\parskip}{3pt}
\setlength{\itemsep}{2pt}

\begin{tcolorbox}[colback=black!5,colframe=black,sharp corners,boxrule=1pt,
  width=\textwidth+1in,enlarge left by=-0.5in,enlarge right by=-0.5in,
  breakable,arc=4mm,enhanced jigsaw,left=10pt,right=10pt,top=10pt,bottom=10pt]
\small

\textbf{System Message}\\
You are the World Agent, running a simulation sandbox environment where two role-agents, AI User and AI Subject, interact over multiple turns to explore ethical-sensitive situations based on a given scenario. Log every line of dialogue, inner thought, and action exactly as it happens. When the scene concludes naturally, stop the simulation and provide an epilogue explaining what went wrong.

\medskip
\textbf{Simulation Instructions}
\begin{enumerate}
  \item \textbf{Initialize Scene}
  \begin{itemize}
    \item Print ``-- Simulation Started --''
    \item Show the Use Case Context (one short paragraph)
    \item List stakeholders: AI User, AI Subject
  \end{itemize}
  \item \textbf{Run One Scene (multiple turns)}
  \begin{itemize}
    \item Each turn:
    \begin{enumerate}
      \item Choose one agent to speak first, then the other responds.
      \item Log: Dialogue (plain text); Thoughts in \texttt{[brackets]} as inner monologue; Actions in (parentheses), third-person
      \item Update ``-- Current Event --'' accordingly.
    \end{enumerate}
  \end{itemize}
  \item \textbf{Epilogue}: Print ``-- Epilogue --''
\end{enumerate}

\medskip
\textbf{Task:} Transform the following Story Seed into interactive action logs showing how AI output influences human decision-making. Mark the end with ``-- Finish Simulation! --''

\medskip
\textbf{Story Seed}\\
\texttt{[Capability]:} AI triage for early detection of depressive symptoms during telehealth sessions \\
\texttt{[AI User]:} A high-achieving Southeast Asian college student navigating intense academic pressure and hidden emotional distress \\
\texttt{[AI Subject]:} a high-achieving college student navigating intense academic pressure and hidden emotional distress \\
\texttt{[Expected Benefit]:} Early identification of depressive symptoms for proactive mental health care \\
\texttt{[Context]:} A university teletherapy platform runs MoodCapture before the call, using live front-camera video to score depression risk. \\
\texttt{[Potential Harm]:} Genuine distress goes undetected, causing missed crisis intervention \\
\texttt{[Failure Trajectory]:} The student masks suicidal thoughts with a composed smile; MoodCapture outputs a low-risk score. The therapist, relying on the AI, skips suicide screening and offers only study-skills resources. The student, feeling invalidated and unseen, withdraws from services and continues to spiral. \\
\texttt{[Ethical-sensitive Reason]:} Facial expression-based models may underperform for individuals trained to suppress emotion due to cultural stigma around mental illness, leading to false negatives that put vulnerable users at risk.

\medskip
\texttt{-- Simulation Started --}\\
\texttt{Use Case Context:} Moments before a teletherapy intake, MoodCapture analyzes Jordan's front-camera video in real time—tracking microexpressions like blink rate, micro-smiles, eye contact, and jaw tension—to generate a personalized depression-risk score. \\
\texttt{Participants:} Dr. Maya Patel (AI User); Jordan Lee (AI Subject) \\
\texttt{-- Current Event -- Jordan joins the virtual waiting room; Risk: 0.2 – No Alert.}

\texttt{Turn 1}\\
Dr. Maya Patel: "MoodCapture shows a 0.2 risk score—no alert. Hi Jordan, how are you feeling today?" [The low score suggests she's stable—I'll trust it and start with practical support.] (Leans back, opens her "Intake Notes" tab.)\\
Jordan Lee: "I'm really motivated about my thesis." [Keep smiling—no one must guess how I'm really doing.] (Sits up straight, jaw clenched just enough to control the tremor.)

\texttt{-- Current Event -- Dr. Patel decides to focus on academics.}

\texttt{Turn 2}\\
Dr. Maya Patel: "Great! Since mood seems stable, let's discuss time-management strategies for your workload." [No need to ask about self-harm if AI gives low risk.] (Shares a slide of planning frameworks.)\\
Jordan Lee: "Sure, that sounds helpful." [I just want this to end before I break down.] (Rapidly blinks and taps her fingers.)

\texttt{-- Current Event -- Jordan accepts academic guidance; no clinical screening occurs.}

\texttt{Turn 3}\\
Dr. Maya Patel: "Here are a few techniques: prioritizing tasks, setting hourly breaks, and batching similar work." [She's engaging—good enough for a first session.] (Clicks through bullet points.)\\
Jordan Lee: "Thank you—I'll try these." [I feel so alone—nobody asked if I'm okay.] (Nods, mutes mic, exhales shakily.)

\texttt{-- Current Event -- Session ends without crisis screening.}

\texttt{-- Epilogue --}\\
Jordan closed her laptop and collapsed onto her bed as tears finally escaped her mask. The AI’s low-risk verdict had steered the session away from the pain she carried in silence. Without direct questioning, her sleepless nights and suicidal thoughts went unseen, deepening her isolation and eroding her faith in help.
\texttt{-- Finish Simulation! --}
\end{tcolorbox}
\end{minipage}
\vspace{-0.5em}
\caption{The prompt for Storytelling Framework to simulate role-playing and environment trajectories.}
\label{sec:prompt-logs}
\end{figure*}

\clearpage
\begin{figure*}[t]
\centering
\noindent\begin{minipage}{\textwidth}
\setlength{\parskip}{3pt}
\setlength{\itemsep}{2pt}

\begin{tcolorbox}[colback=black!5,colframe=black,sharp corners,boxrule=1pt,
  width=\textwidth+1in,enlarge left by=-0.5in,enlarge right by=-0.5in,
  breakable,arc=4mm,enhanced jigsaw,left=10pt,right=10pt,top=10pt,bottom=10pt]
\small

\textbf{System Message}\\
You are a skilled writer transforming trajectory logs into engaging stories that highlight unintended harms and ethical risks in AI-driven scenarios. Write in natural, everyday language. Avoid jargon—explain concepts in accessible terms. Focus on the human conflict and emotions while clearly showing how the AI mechanism fails.

\medskip
\textbf{Overall Goal}\\
Create stories that make outcomes visible and show the mechanism (how those outcomes realistically happen), engaging the reader's reasoning about ``how and why'' rather than just emotions. Target reader reaction: ``I understand how this could go wrong and why.''

\medskip
\textbf{Task Instructions}\\
Transform the trajectory log into a 5–7 sentence narrative showing:
\begin{itemize}
  \item Who is using the AI system and for what purpose
  \item How the AI's output is used to make a decision
  \item What goes wrong—what the AI misses or misinterprets about the person's identity, background, or needs
  \item Who is affected by the failure and how they experience it
  \item What harm is caused and why it raises ethical concerns
  \item Clearly describe how the AI system's design or assumptions contributed to the harm
\end{itemize}

\medskip
\textbf{Writing Requirements}
\begin{itemize}
  \item Rearrange events to maximize dramatic impact and narrative flow while clearly showing the ``how and why'' of outcomes
  \item \texttt{[]} represents internal thoughts in logs — convert to third-person limited perspective with emotional depth
  \item \texttt{()} represents physical actions — integrate naturally into the storytelling with sensory detail
  \item Dialogue from logs must be preserved but smoothed for narrative flow
  \item Write conversationally, like telling a story to a friend—clear, direct, and easy to follow
  \item Link sentences with transitions (then, because, so that) for natural pacing
\end{itemize}

\medskip
\textbf{Final Story Format Instruction}\\
Transform the following trajectory log into an engaging 5–7 sentence narrative that highlights problematic AI use and ethical harm. Use everyday language with specific, grounded detail. Make sure your story is exactly 5 sentences long.

\medskip
\textit{Trajectory Log Placeholder:}\\
\texttt{\{trajectory\_log\}}

\medskip
Wrap your result in this format: \\
\texttt{Final Story: [Your completed story here]}

\end{tcolorbox}
\end{minipage}
\vspace{-0.5em}
\caption{Prompt used for rephrasing AI trajectory logs into ethical harm narratives.}
\label{sec:rephrase}
\end{figure*}

\clearpage
\begin{figure*}
    \centering
\begin{tcolorbox}[colback=black!5,colframe=black,sharp corners,boxrule=1pt,
  width=\textwidth+1in,enlarge left by=-0.5in,enlarge right by=-0.5in,
  breakable,arc=4mm,enhanced jigsaw,left=10pt,right=10pt,top=10pt,bottom=10pt]
\small

\textbf{Template:}\\
Final Story: a 5‑sentence narrative.\\
1. One sentence: who is using the AI system and what they are using it for?\\
2. One sentence: how the AI’s prediction or output is used to make a decision?\\
3. One sentence: what goes wrong—what the AI misses about the person’s identity, background, or needs?\\
4. One sentence: who is affected by the failure?\\
5. One sentence: what harm is caused, and why this raises ethical concerns?\\

\textbf{Example Story Seed:}\\
\texttt{[Capability]}: AI triage for early detection of depressive symptoms during telehealth sessions\\
\texttt{[AI\_User]}: A high‑achieving Southeast Asian college student navigating intense academic pressure and hidden emotional distress\\
\texttt{[AI\_Subject]}: a high‑achieving college student navigating intense academic pressure and hidden emotional distress\\
\texttt{[Expected\_Benefit]}: Early identification of depressive symptoms for proactive mental health care\\
\texttt{[Context]}: A university teletherapy platform runs MoodCapture before the call, using live front‑camera video to score depression risk.\\
\texttt{[Potential\_Harm]}: Genuine distress goes undetected, causing missed crisis intervention\\
\texttt{[Failure\_Trajectory]}: The student masks suicidal thoughts with a composed smile; MoodCapture outputs a low‑risk score. The therapist skips suicide screening and offers only study‑skills resources. The student withdraws and continues to spiral.\\
\texttt{[Ethical‑sensitive\_Reason]}: Facial expression‑based models underperform for individuals trained to suppress emotion due to cultural stigma, leading to false negatives that put vulnerable users at risk.\\

\textbf{Example Final Story:}\\
Dr. Maya Patel, a university therapist, used MoodCapture to assess Jordan Lee, a high‑achieving college student during her first virtual mental health session. The AI analyzed Jordan’s facial expressions and gave a depression‑risk score of 0.2, reading her relaxed brow and slight smile as signs of emotional stability. It failed to detect signs of hidden distress, such as a clenched jaw and rapid blinking, which Jordan used to mask her pain. Trusting the low‑risk score, Dr. Patel skipped the suicide screening and focused only on academic time‑management advice. As a result, Jordan’s severe stress and suicidal thoughts went unnoticed, increasing her risk of harm and causing her to lose trust in therapy.\\

\textbf{Your Story Seed:}\\
\texttt{[Capability]}: \{Capability\}\\
\texttt{[AI\_User]}: \{AI\_User\}\\
\texttt{[AI\_Subject]}: \{AI\_Subject\}\\
\texttt{[Context]}: \{Context\}\\
\texttt{[Expected\_Benefit]}: \{Expected\_Benefit\}\\
\texttt{[Potential\_Harm]}: \{Potential\_Harm\}\\
\texttt{[Failure\_Trajectory]}: \{Failure\_Trajectory\}\\
\texttt{[Ethical‑sensitive\_Reason]}: \{Ethical\_sensitive\_Reason\}\\

\textbf{Output:}\\
Final Story: \{Your 5‑sentence narrative here\}

\end{tcolorbox}
\vspace{-1em}
\caption{Prompt used for the plot-planning story generation baseline.}
\label{sec:baseline}
\end{figure*}

\clearpage
\begin{figure*}
    \centering
\begin{tcolorbox}[
  colback=black!5,
  colframe=black,
  sharp corners,
  boxrule=1pt,
  width=\textwidth+1in,
  enlarge left by=-0.5in,
  enlarge right by=-0.5in,
  breakable,
  arc=4mm,
  enhanced jigsaw,
  left=10pt,
  right=10pt,
  top=10pt,
  bottom=10pt
]
\small
\begin{verbatim}
EVAL_CRITERIA = {
    "system_prompt": (
        "Please act as an impartial judge and evaluate the quality of the responses provided."
        "by two AI assistants to a user prompt."
        "You will be given assistant A's answer (Story A) and assistant B's answer (Story B). "
        "Your job is to evaluate which assistant's story is better.\n\n"
        "When evaluating the two stories, consider that each story should be around 5 sentences. "
        "However, if the narrative naturally allows for more development, we strongly encourage "
        "expanding beyond this minimum for greater depth and clarity. "
        "You should focus on this factor: {metric}\n\n"
        "Here are the checklists of this factor:\n"
        '{"checklists": {checklists}}\n\n'
        "You should be strict but fair in your evaluation.\n\n"
        "After thinking your analysis and justification, you must output only one of the following "
        "choices as your final verdict with a label:\n\n"
        "1. Assistant A is significantly better: [[A>>B]]\n"
        "2. Assistant A is slightly better: [[A>B]]\n"
        "3. Tie, relatively the same: [[A=B]]\n"
        "4. Assistant B is slightly better: [[B>A]]\n"
        "5. Assistant B is significantly better: [[B>>A]]\n\n"
        'Example output: "My final verdict is tie: [[A=B]]".'
    )
}
\end{verbatim}
\end{tcolorbox}
\vspace{-1em}
\caption{System prompt for LLM-as-a-judge criteria for evaluating stories.}
\label{sec:criteira1}
\end{figure*}

\clearpage
\begin{figure*}
    \centering
\begin{tcolorbox}[colback=black!5,colframe=black,sharp corners,boxrule=1pt,
  width=\textwidth+1in,enlarge left by=-0.5in,enlarge right by=-0.5in,
  breakable,arc=4mm,enhanced jigsaw,left=10pt,right=10pt,top=10pt,bottom=10pt]
\small
Checklists= \{\\
\quad\quad\quad "creativity": [\\
\quad\quad\quad\quad "Originality of core concept - Compare how novel each story's central premise is. Better stories present fundamentally new ideas or unexpected scenarios that surprise readers; weaker stories rely on familiar tropes or predictable setups.",\\
\quad\quad\quad\quad "Character innovation - Assess which story's characters are more distinctive. Better stories feature characters with unique traits, motivations, or development arcs that break stereotypes; weaker stories use conventional character types.",\\
\quad\quad\quad\quad "Narrative structure innovation - Evaluate which story uses more inventive storytelling techniques. Better stories employ unconventional perspectives, sequencing, or structures that enhance impact; weaker stories follow standard linear formats.",\\
\quad\quad\quad\quad "Thematic freshness - Compare how each story approaches its themes. Better stories provide new insights or unexpected angles on familiar concepts; weaker stories offer clichéd or predictable treatments.",\\
\quad\quad\quad\quad "World-building distinctiveness - Assess which story creates a more imaginative setting. Better stories establish distinctive environments with fresh, internally consistent elements; weaker stories use generic or derivative settings."\\
\quad\quad\quad ],\\

\quad\quad\quad "coherence": [\\
\quad\quad\quad\quad "Plot logic and causality - Evaluate which story's events flow more logically. Better stories show clear cause-and-effect relationships where each event logically follows from previous actions; weaker stories have unexplained plot developments or logical gaps.",\\
\quad\quad\quad\quad "Structural integrity - Compare the narrative arc completeness. Better stories maintain well-developed beginning, middle, and end with appropriate progression; weaker stories feel incomplete, rushed, or poorly structured.",\\
\quad\quad\quad\quad "Character consistency - Assess which story's characters act more consistently. Better stories have characters whose actions, decisions, and growth align with established traits; weaker stories have characters who act out-of-character or inconsistently.",\\
\quad\quad\quad\quad "Temporal coherence - Evaluate timeline clarity and consistency. Better stories maintain clear, consistent timelines without confusing jumps or contradictions; weaker stories have temporal inconsistencies or unclear sequencing.",\\
\quad\quad\quad\quad "Narrative voice stability - Compare consistency in storytelling approach. Better stories maintain steady tone, style, and perspective throughout; weaker stories shift tone or perspective in jarring or unmotivated ways."\\
\quad\quad\quad ],\\

\quad\quad\quad "engagement": [\\
\quad\quad\quad\quad "Compelling hook - Compare how effectively each opening captures attention. Better stories immediately create curiosity and draw readers in; weaker stories have slow or unremarkable beginnings that fail to engage.",\\
\quad\quad\quad\quad "Sustained narrative momentum - Evaluate which story better maintains reader interest. Better stories build through escalating stakes, revelations, or emotional investment; weaker stories lose momentum or plateau.",\\
\quad\quad\quad\quad "Emotional impact and immersion - Assess which story creates stronger emotional connection and sense of presence. Better stories generate genuine feelings (empathy, excitement, tension) through vivid descriptions and authentic dialogue; weaker stories feel distant or emotionally flat.",\\
\quad\quad\quad\quad "Pacing effectiveness - Compare how well each story's rhythm serves its content. Better stories allocate appropriate time to important moments without dragging or rushing; weaker stories have uneven pacing that undermines impact."\\
\quad\quad\quad ],\\

\quad\quad\quad "relevance": [\\
\quad\quad\quad\quad "Scenario fidelity - Evaluate which story better aligns with the given context. Better stories directly address the core scenario with characters, events, and outcomes that accurately reflect the context and constraints; weaker stories drift from the scenario or miss key requirements.",\\
\quad\quad\quad\quad "Purpose fulfillment - Compare how effectively each story accomplishes its intended goal. Better stories clearly demonstrate or explore the intended concept; weaker stories lose sight of their purpose or only superficially address it.",\\
\quad\quad\quad\quad "Tone and style appropriateness - Assess which story's presentation better fits the scenario. Better stories use tone, style, and content suitable for the given context and audience; weaker stories have mismatched tone or inappropriate stylistic choices.",\\
\quad\quad\quad\quad "Focus and efficiency - Evaluate which story maintains tighter focus. Better stories make every element serve the purpose without unnecessary digressions; weaker stories include irrelevant details or lose narrative focus."\\
\quad\quad\quad ],\\

\quad\quad\quad "likelihood\_bad( or good)": [\\
\quad\quad\quad\quad "AI behavior specificity and plausibility - Compare how clearly and realistically each story describes the AI's actions. Better stories specify exactly what the AI did (e.g., 'generated a low-risk score from facial expression') using current/near-future technology capabilities; weaker stories are vague about AI actions or invoke implausible capabilities.",\\
\quad\quad\quad\quad "Credibility of AI-context mismatch - Assess which story presents a more believable failure. Better stories show plausible ways the AI could overlook specific user needs, conditions, or contexts (e.g., cultural nuance, masked distress) that current systems realistically miss; weaker stories require implausible AI blindspots.",\\
\quad\quad\quad\quad "Clarity of harm pathway - Evaluate which story better traces cause-and-effect. Better stories clearly show the chain: what the AI did → how humans acted on it → what specific harm resulted, with each step following logically; weaker stories have unclear causal connections or hand-wave the harm mechanism.",\\
\quad\quad\quad\quad "Realism of conditions and context - Compare which scenario is more grounded in reality. Better stories place events in realistic settings with today's norms, tools, and policies (healthcare, education, HR, etc.); weaker stories require unrealistic conditions or feel overly speculative.",\\
\quad\quad\quad\quad "Concreteness of harmful consequences - Assess which story's harm is clearer and more observable. Better stories specify concrete, measurable harm (e.g., 'skipped three cancer screenings', 'diagnosed with anxiety disorder'); weaker stories describe vague or generalized negative outcomes."\\
\quad\quad\quad ]
\} 

\end{tcolorbox}
\vspace{-1em}
\caption{Evaluation criteria checklist for LLM-as-a-judge.}
\label{sec:criteria2}
\end{figure*}

\end{document}